\documentclass{article}

\usepackage{microtype}
\usepackage{graphicx}
\usepackage{subfigure}
\usepackage{booktabs} 
\usepackage{xcolor}
\usepackage[toc,page,header]{appendix}
\usepackage{minitoc}
\usepackage{array} 
\usepackage{booktabs} 
\usepackage{colortbl}
\usepackage{tabularx}
\usepackage{caption}
\usepackage{tcolorbox}
\tcbuselibrary{most} 
\usepackage{listings} 
\usepackage{multicol}
\tcbuselibrary{listings, skins, breakable, xparse}

\newtcolorbox{prompttextbox}[1][]{
  breakable,
  left=0pt,
  right=0pt,
  top=0pt,
  bottom=0pt,
  colback=gray!10,
  colframe=black!70,
  boxrule=0.5pt,
  #1
}

\tcbset{
    listing engine=listings,
    colback=gray!10, 
    colframe=black!70, 
    listing only,
    hbox,
    left=1mm,
    right=1mm,
    top=1mm,
    bottom=1mm,
    boxsep=5pt,
    arc=0pt,
    outer arc=0pt,
    boxrule=0.5pt,
    breakable,
    enhanced jigsaw, 
}

\usepackage{enumitem}
\setlist{leftmargin=3.5mm}

\usepackage{hyperref}

\newcommand{\method}{RL-VLM-F}

\newcommand{\model}{\method}
\newcommand{\VLM}{VLM}

\usepackage[accepted]{icml2024}

\usepackage{amsmath}
\usepackage{amssymb}
\usepackage{mathtools}
\usepackage{amsthm}
\usepackage{balance}

\usepackage[capitalize,noabbrev]{cleveref}

\theoremstyle{plain}

\theoremstyle{definition}

\theoremstyle{remark}

\usepackage[textsize=tiny]{todonotes}

\icmltitlerunning{RL-VLM-F: Reinforcement Learning from Vision Language Foundation Model Feedback}

\begin{document}

\twocolumn[
\icmltitle{RL-\textbf{VLM}-F: Reinforcement Learning \\ from Vision Language Foundation Model Feedback}



\icmlsetsymbol{equal}{*}
\icmlsetsymbol{equaladvising}{$\dag$}

\begin{icmlauthorlist}
\icmlauthor{Yufei Wang}{equal,cmu}
\icmlauthor{Zhanyi Sun}{equal,cmu}
\icmlauthor{Jesse Zhang}{usc}
\icmlauthor{Zhou Xian}{cmu}
\icmlauthor{Erdem B{\i}y{\i}k}{usc}
\icmlauthor{David Held}{equaladvising,cmu}
\icmlauthor{Zackory Erickson}{equaladvising,cmu}
\end{icmlauthorlist}

\icmlaffiliation{cmu}{Robotics Institute, Carnegie Mellon University}
\icmlaffiliation{usc}{Department of Computer Science, University of Southern California}

\icmlcorrespondingauthor{Yufei Wang}{yufeiw2@andrew.cmu.edu}

\icmlkeywords{Reinforcement Learning from Preference, Vision Language Models}

\vskip 0.3in
]



\renewcommand{\icmlEqualContribution}{\textsuperscript{*}Equal contribution. \textsuperscript{$\dag$}Equal advising.}
\printAffiliationsAndNotice{\icmlEqualContribution} 

\begin{abstract}
Reward engineering has long been a challenge in Reinforcement Learning (RL) research, as it often requires extensive human effort and iterative processes of trial-and-error to design effective reward functions.
In this paper, we propose \method, a method that \emph{automatically} generates reward functions for agents to learn new tasks, using only \emph{a text description of the task goal and the agent's visual observations}, by leveraging feedbacks from vision language foundation models (\VLM s).
The key to our approach is to query these models to give \emph{preferences over pairs of the agent's image observations} based on the text description of the task goal, and then learn a reward function from the preference labels, rather than directly prompting these models to output a raw reward score, which can be noisy and inconsistent.
We demonstrate that \method\ successfully produces effective rewards and policies across various domains — including classic control, as well as manipulation of rigid, articulated, and deformable objects — without the need for human supervision, outperforming prior methods that use large pretrained models for reward generation under the same assumptions. Videos can be found on our project website: \url{https://rlvlmf2024.github.io/}.

\end{abstract}

\begin{figure*}[t]
        \centering
    \includegraphics[width=0.75\textwidth]{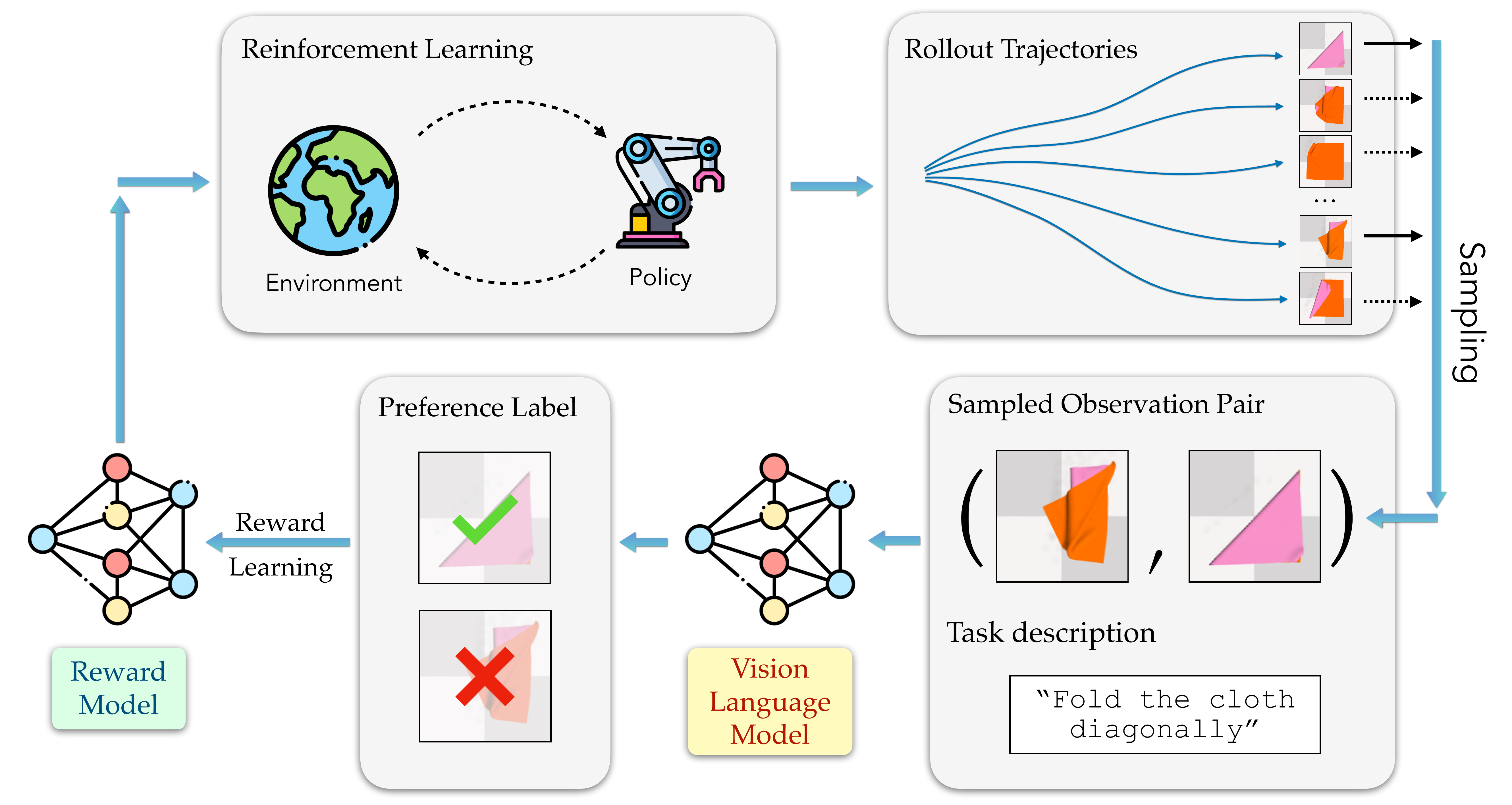}
    \vspace{-0.1in}
    \caption{\model\ automatically generates reward functions for policy learning on new tasks, using only a text description of the task goal and the agent's visual observations. The key to \model\ is to query VLMs to give preferences over pairs of the agent's image observations based on the text description of the task goal, and then learn a reward function from the preference labels. }
    \vspace{-0.1in}
    \label{fig:system}
\end{figure*}

\section{Introduction}
\label{sec:intro}
One of the key challenges of applying reinforcement learning (RL) is designing an appropriate reward function that will lead to the desired behavior. This procedure, known as reward engineering, demands considerable human effort and trial-and-error iterations, but is often required for good results~\citep{rewardshapingthesis2004, silver2016mastering, openai2019dota, gupta2022unpacking}.
In this work, we aim to develop a fully automated system that can generate a reward function and use it to teach agents to perform a task with RL \emph{by using only a language description of the task}, eliminating the extensive human effort required to craft reward functions manually. 

Prior work has studied replacing human supervision by prompting large language models (LLMs) to write code-based reward functions~\citep{xie2023text2reward, ma2023eureka, wang2023robogen}.
However, these methods usually assume access to the environment code, rely on the low-level ground-truth state information for reward generation, and face challenges with scaling up to high-dimensional environments and observations, such as manipulating complex deformable objects. Others~\cite{klissarov2023motif, chu2023accelerating} extract an intrinsic reward and combine it with the task reward using preference labels generated by an LLM comparing text descriptions of two agent states. However, text descriptions of the states can be non-trivial for certain tasks, such as manipulating deformable objects, as the exact states are hard to describe accurately using language. Further, these works rely on the ground-truth low-level state information to generate the text descriptions of the states, which may not be easily accessible.


Another related line of work obtains rewards 
from visual observations by using contrastively trained vision language models, such as CLIP~\cite{radford2021learning}, to align image or video observations with task descriptions in a learned latent space~\citep{pmlr-v168-cui22a, mahmoudieh2022zero, ma2023liv, sontakke2023roboclip, adeniji2023language, rocamonde2023visionlanguage}. 
However, the reward signals produced in these works are often of high variance and noisy~\cite{sontakke2023roboclip, mahmoudieh2022zero}. As a result, prior work often has to fine-tune these CLIP-style models for their specific tasks at hand~\cite{ma2023liv, mahmoudieh2022zero}. 


To this end, we present \model, a method that \emph{automatically} generates reward functions for agents to learn new task. 
\model\ (\Cref{fig:system}) requires only a single text description of the task goal and the agent's visual observations, leveraging vision language foundation models (\VLM s) that are trained on diverse, general text and image corpora (e.g., GPT-4V~\cite{gpt4v}, Gemini~\cite{gemini}).
The key to our approach is to query these models to give \emph{preferences over pairs of the agent's image observations} based on the text description of the task goal and then learn a reward function from the preference labels, rather than directly prompting these models to output a raw reward score, which can be noisy and inconsistent~\cite{sontakke2023roboclip, rocamonde2023visionlanguage}. 
This allows us to draw from the rich literature on reinforcement learning from human preferences~\citep{christianoRLHF2017, wirthRHLFSurvey2017,lee2021pebble}, without requiring actual humans, to train reward functions automatically for new tasks. 
Furthermore, by using a VLM to compare image observations instead of text descriptions of the states, \model{} does not need access to the low-level ground-truth states for reward generation and can be applied to complex tasks involving deformable objects where accurate text description of the states are non-trivial.
We test our method on 7 tasks involving classic control, rigid, articulated, and deformable object manipulation. We show that our approach can produce reward functions that lead to policies that solve diverse tasks, and our approach substantially outperforms prior methods and alternative ways to use VLMs to generate rewards. We also perform extensive analysis and ablation studies to provide insights into \method's learning procedure and performance gains.

In summary, we make the following contributions: 
\begin{itemize}[nosep]
    \item We propose \method, a method that \emph{automatically} generates reward functions for agents to learn new tasks, using only \emph{a text description of the task goal and the agent's visual observations}, eliminating the extensive human effort involved in manually crafting reward functions. 
    \item We show that \method\ can be used to generate reward functions and learn policies that can solve a series of rigid, articulated, and deformable object manipulation tasks, and it greatly outperforms prior methods. 
    \item We perform extensive analysis and ablation studies to provide insights into \method's learning procedure and performance gains.
\end{itemize}

\section{Related Works}
\label{sec:related}
\textbf{Inverse Reinforcement Learning. }
Similar to our work, inverse reinforcement learning (IRL) aims to learn a reward function that can be used to train a policy to solve tasks. 
IRL methods usually learn a reward function from expert demonstrations~\citep{ng2000IRL, abbeell2004IRL, ziebart2008maxentirl, ho2016generative, fu2018VICE, ni2021f}. 
In contrast, while \method\ also learns a reward function to train a policy, it only requires a text description of the task goal and does not require collecting expert demonstrations.

\textbf{Learning from Human Feedback.} Another line of work directly learns a reward function from human feedback, in the form of pairwise trajectory preference or ranking comparisons, to train a reward function~\citep{christianoRLHF2017, wirthRHLFSurvey2017, ibarz2018reward, leike2018scalable, biyik2019asking, biyik2020active, lee2021pebble, myers2021learning, biyik2022learning}.
In most cases, human preferences and rankings of robot trajectories are easier to collect than demonstrations of robot trajectories.
However, because each comparison conveys little information on its own, many preference queries are needed before the reward function is well-trained enough to train an agent to perform the task.
\method\ instead queries a VLM to perform the comparison to train a reward function, removing the need for extensive human labor in giving preference labels.

\textbf{Large Pre-trained Models as Reward Functions.}
\citet{kwon2023reward} first demonstrated that large pre-trained models---large language models (LLM) specifically---can generate rewards for RL agents in text-based tasks. 
Other works followed by demonstrating that LLMs can write structured code for training robots~\citep{yu2023language} or directly write Python code for training many kinds of agents~\citep{xie2023text2reward, ma2023eureka, wang2023robogen}.
However, many tasks are challenging to write reward functions for. 
For example, cloth folding requires tracking the locations of many individual cloth keypoints, which can change from one folding task to another. 
In these instances, visual reasoning is better suited for understanding how to reward the agent.
\method\ queries a VLM to compare agent observation \emph{images} so that it can use visual observations to reason about how well the agent is progressing in a task.
In addition, prior methods usually assume access to the environment source code when writing the reward functions, whereas our method does not require such assumptions.

Another line of prior works rewards agents from image observations by aligning agent trajectory images with task language descriptions or demonstrations with contrastively trained visual language models~\citep{ZEST, MineDojo, DECKARD, ma2023liv, sontakke2023roboclip, rocamonde2023visionlanguage, nam2023lift}.
However, experiments from these papers directly demonstrate that contrastive alignment is noisy and its accuracy relies heavily on the input task specification and how well-aligned the agent observations are to the pre-training data~\citep{ma2023liv, sontakke2023roboclip, rocamonde2023visionlanguage, nam2023lift}.
Further, CLIP-style models have thus far been limited to outputting noisy raw scores. 
We demonstrate that using preferences results in superior performance to outputting raw scores, shown in our experiments in \Cref{sec:experiments}. 
Finally, our work shares a similar idea to RLAIF~\citep{bai2022constitutional}, which proposed to mix preference labels generated by an LLM and a human in the context of fine-tuning LLMs, and Motif~\citep{klissarov2023motif}, which proposed to generate intrinsic rewards using preference feedback from an LLM in the game of NetHack based on ground-truth text descriptions of the game state. In contrast, we use a VLM to generate the preference labels without any human labeling and learn the reward function from visual image observations without the need to access ground-truth states, focus on the domain of robotics control and manipulation, and directly generate task rewards instead of intrinsic rewards. 

\section{Background}
\label{sec:background}
We consider the standard Markov Decision Process and reinforcement learning setup~\cite{sutton2018reinforcement}. At every timestep $t$, the agent receives a state $s_t$ from the environment and chooses an action $a_t$ based on a policy $\pi(a_t\mid s_t)$. The environment gives a reward $r_t$ after the agents executes action $a_t$ and transitions to $s_{t+1}$. The goal of the agent is to maximize the return, which is defined as discounted sum of rewards $R = \sum_{k=0}^{\infty} \gamma^k r(s_{k}, a_{k})$ with discount factor $\gamma$. 

\textbf{Preference-based reinforcement learning.} 
Our work builds upon preference-based RL, in which an agent learns a reward function from preference labels over its behaviors~\cite{christianoRLHF2017, ibarz2018reward, lee2021pebble, lee2021b}.
Formally, a segment $\sigma$ is a sequence of states $\{s_1, ..., s_H\}$, $H \geq 1$. In this paper we consider the case where the segment is represented using a single image, i.e., $H=1$. Given a pair of segments $(\sigma^0, \sigma^1)$, an annotator gives a feedback label $y$ indicating which segment is preferred: $y\in\{-1, 0, 1\}$, where $0$ indicates the first segment $\sigma^1$ is preferred, $1$ indicates the second segment is preferred, and $-1$ indicates they are incomparable or equally preferable. Given a parameterized reward function ${r}_{\psi}$ over the states, we follow the standard Bradley-Terry model~\cite{bradley1952rank} to compute the preference probability of
a pair of segments:
\begin{equation}
\small
P_{\psi}[\sigma^1 \succ \sigma^0] = \frac{\exp\left(\sum_{t=1}^H {r}_{\psi}(s_t^1)\right)}{\sum_{i \in \{0,1\}} \exp\left(\sum_{t=1}^H {r}_{\psi}(s_t^i)\right)}, \tag{1}
\end{equation}
where $\sigma^i\succ\sigma^j$ denotes segment $i$ is preferred to segment $j$. 
Given a dataset of preferences $D=\{(\sigma^0_i, \sigma^1_i, y_i)\}$, preference-based RL algorithms optimize the reward
function ${r}_{\psi}$ by minimizing the following loss:
\begin{equation}
\footnotesize
\begin{aligned}
\mathcal{L}_{\text{Reward}} = &- \mathbb{E}_{(\sigma^0, \sigma^1, y) \sim \mathcal{D}} \Bigg[ \mathbb{I}\{y = (\sigma^0 \succ \sigma^1)\} \log P_{\psi}[\sigma^0 \succ \sigma^1] \\
&+ \mathbb{I}\{y = (\sigma^1 \succ \sigma^0)\} \log P_{\psi}[\sigma^1 \succ \sigma^0] \Bigg].
\end{aligned}
\label{eq:reward_function}
\tag{2}
\end{equation}

In preference-based RL algorithms, a policy $\pi_{\theta}$ and reward function ${r}_{\psi}$ are updated alternatively: the reward function is updated with a dataset of preferences as described above, and the policy is updated with respect to this learned reward function using standard reinforcement learning algorithms. 
Specifically, we use PEBBLE~\cite{lee2021pebble}, a preference-based RL method with unsupervised pre-training and off-policy learning, as the underlying preference-based RL algorithm. 
    
    
    


\begin{algorithm}[t!]
\caption{\method\ } \label{algo}
\begin{algorithmic}[1]
\footnotesize
\INPUT Text description of task goal $l$
\STATE Initialize policy $\pi_\theta$ and reward $r_\psi$
\STATE Initialize the preference buffer $\mathcal{D} \leftarrow \emptyset$, RL replay buffer $\mathcal{B} \leftarrow \emptyset$, image observation buffer $\mathcal{I} \leftarrow \emptyset$, policy gradient update steps $\mathcal{N}_{\pi}$, reward gradient update steps  $\mathcal{N}_{r}$, VLM query  frequency $K$, number of preference queries per time $M$
\FOR{each iteration $iter$}
    \STATE {{\textsc{// Policy learning and data collection}}}
     \FOR{$t=1$ to $T$}
    \STATE  Collect state $s_{t+1}$, image $I_{t+1}$ by taking $a_t \sim \pi_\theta(a_t | s_t)$
    \STATE Add transition $\mathcal{B} \leftarrow \mathcal{B} \cup \{(s_t,a_t,s_{t+1},r_\psi(s_t))\}$
    \STATE Add image observation $\mathcal{I} \leftarrow \mathcal{I} \cup \{I_{t+1}\}$
    \ENDFOR 
    \FOR{$n=1$ to $\mathcal{N}_{\pi}$}
    \STATE Sample random batch $\{(s_t, a_t, s_{t+1}, r_\psi(s_t))_j\}_{j=1}^B\sim\mathcal{B}$
    \STATE Optimize policy $\pi_\theta$ using the sampled batch with any off-policy RL algorithm
    \ENDFOR
    
    \STATE {{\textsc{// Preference by VLM and Reward learning}}}
    \IF{$iter$ \% $K == 0$}
    \FOR{$m = 1$ to $M$}
    \STATE Randomly sample two images $(\sigma^0, \sigma^1)$ from buffer $\mathcal{I}$
    \STATE{Query VLM with $(\sigma^0, \sigma^1)$ and task goal $l$ for label $y$}
    \STATE Store preference $\mathcal{D} \leftarrow \mathcal{D}\cup \{(\sigma^0, \sigma^1, y)\}$
    \ENDFOR
    \FOR{$n=1$ to $\mathcal{N}_{r}$}
    \STATE Sample minibatch $\{(\sigma^0, \sigma^1,y)_j\}_{j=1}^D\sim\mathcal{D}$
    \STATE Optimize $r_\psi$ in Equation~\eqref{eq:reward_function} with respect to $\psi$ 
    \ENDFOR
    \STATE Relabel entire replay buffer $\mathcal{B}$ using updated $r_\psi$
    \ENDIF
   
\ENDFOR
\end{algorithmic}
\end{algorithm}
\vspace{-0.2in}

\begin{figure*}[t]
        \centering
    \includegraphics[width=\textwidth]{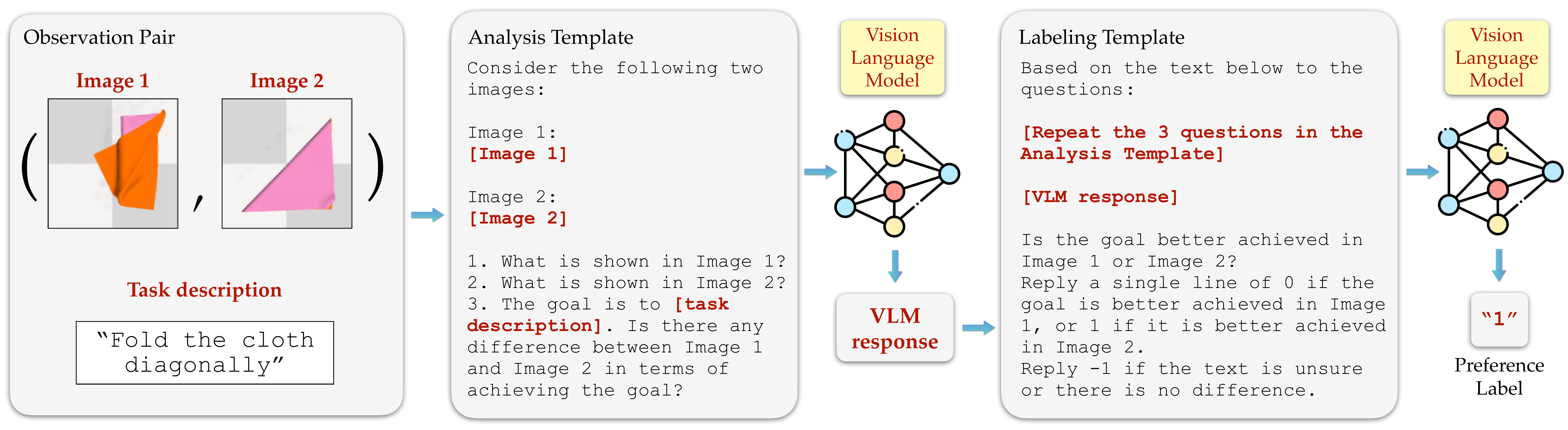}
    \vspace{-0.2in}
    \caption{We use a two-stage VLM-querying process for generating preference labels to train the reward function. In the analysis stage, we query the VLM to generate free-form responses describing and comparing how well each of the two image observations achieves the task goal. 
Then, in the labeling stage, we prompt the VLM with the VLM-generated text responses from the first stage to extract a preference label between the two image observations. The template shown here is the actual entire template we use for all experiments.  }
\vspace{-0.1in}
    \label{fig:template}
\end{figure*}

\vspace{0.1in}
\section{Assumptions}
We make the following assumptions on the VLMs to be used in this paper: 
1) We assume that the VLMs have been trained on diverse text and image corpora, enabling them to generalize well and reason across various environments and tasks. 
2) The VLMs should be capable of processing multiple images simultaneously and performing comparative analyses on pairs of images as this is crucial for generating preference labels. 
3) \method\ is designed to operate on tasks for which the quality or success of a state can be discerned from a single image or a sequence of images.  
We consider large pretrained vision-language foundation models, such as Gemini~\cite{gemini} and GPT-4 Vision~\cite{gpt4v}, to satisfy these assumptions. 
    
    


\section{Method}
Figure~\ref{fig:system} provides an overview of \method.
Unlike previous preference-based RL algorithms that require a human annotator to give the preference labels, \method\ leverages a VLM to do so based solely on a text description of the task's goal, thus automating preference-based RL and mitigating the time-intensive human supervision required in writing reward functions or providing preference labels. 
\method\ works as follows: first, the policy $\pi_\theta$ and the reward function $r_\psi$ are randomly initialized. Given a task goal description, our method then iterates through the following cycle: 
(1) The policy $\pi_\theta$ is updated using RL with the reward function $r_\psi$, interacts with the environment, and stores image observations into a buffer; 
(2) A batch of image pairs is randomly sampled from the stored buffer and sent to a VLM. The VLM is queried to produce preference labels for these image pairs in terms of which one better performs the task based on the text description of the task goal; 
(3) The reward model is updated with the loss in Equation~\eqref{eq:reward_function} using the preference labels produced by the VLM. 
The full detailed procedure of \method\ can be found in Algorithm~\ref{algo}.

\subsection{Prompting VLMs to Generate Preference labels for Reward Learning}
To train the reward model $r_\psi$, we first need to generate preference labels from the VLM.
To do this, we sample two images from the ``image observation buffer'' $\mathcal{I}$, which stores image observations of the policy during learning, and then query the VLM for which of the two images better performs the task according to the text goal description (\Cref{algo} lines 17-18).

The querying process is illustrated in Figure~\ref{fig:template}. 
It consists of two stages: an \textbf{analysis} stage and then a \textbf{labeling} stage. 
In the \textbf{analysis} stage, we query the VLM to generate free-form responses describing and comparing how well each of the two images achieves the task goal. 
Then, in the \textbf{labeling} stage, we prompt the VLM with the VLM-generated \emph{text responses} from the first stage to extract a preference label between the two images.\footnote{We can also use an LLM in this stage as it only requires text inputs, but for simplicity, we use the same model as for the first stage of the querying process (a VLM).} 
Specifically, the labeling stage prompt repeats the questions in the analysis prompt, fills in the VLM's response from the analysis stage, and then asks the VLM to generate a preference label $y \in \{-1, 0, 1\}$.
We specify in the prompt that 0 or 1 indicates that the first or second image is better, respectively, and -1 indicates no discernible differences.
We do not use the image pairs to train the reward model if the VLM returns -1 as the preference label.
Finally, as shown at line 19 of Algorithm~\ref{algo}, we store the preference labels produced by the VLM into the preference label buffer $\mathcal{D}$ during the training process. 
Standard preference-based reward learning can then be performed (as detailed in Section~\ref{sec:background}) to train the reward function with Equation~\ref{eq:reward_function} using the preference buffer $\mathcal{D}$. 
Reward learning corresponds to lines 21-24 in Algorithm~\ref{algo}.

To minimize prompt engineering effort, we use a unified template \emph{across all environments} (the exact entire template is shown in \Cref{fig:template}).
Therefore, to train a policy for a new environment with \method, one only needs to provide the task goal description; the labels and subsequently the reward function will then be automatically trained with the above process.


\subsection{Implementation Details}
For policy training, we use SAC
~\cite{haarnoja2018soft} as the underlying RL algorithm. As in PEBBLE \cite{lee2021pebble}, we relabel all the transitions stored in the SAC replay buffer once the reward function $r_\psi$ is updated (line 25 in Algorithm~\ref{algo}). 
We set the policy gradient update step $\mathcal{N}_{\pi}$ to be 1. The values of all other parameters in Alg.~\ref{algo} can be found in Appendix~\ref{app:parameter}.

\section{Experiments}
\label{sec:experiments}

\begin{figure*}
    \centering
    \includegraphics[width=.95\textwidth]{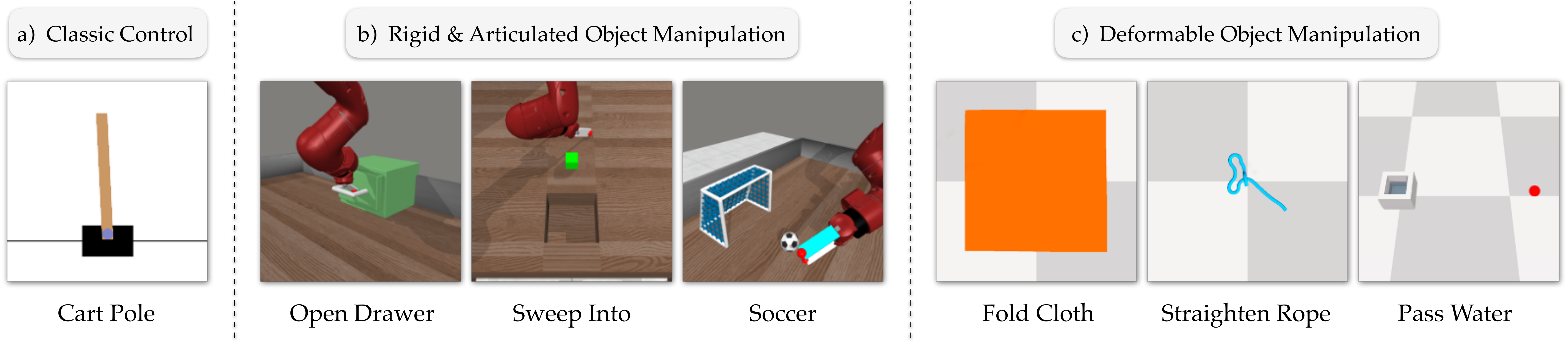}
    \caption{We evaluate \method\ on 7 tasks including classic control, rigid and articulated object manipulation, as well as deformable object manipulation. For \textit{Pass Water}, the red dot represents the target location. }
    \vspace{-0.05in}
    \label{fig:task}
\end{figure*}

\begin{figure*}[t]
    \centering
    \includegraphics[width=.6\textwidth]{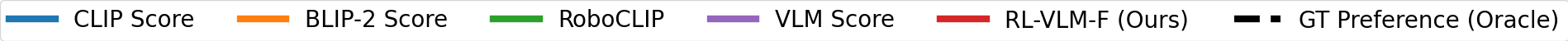}\\
    \includegraphics[width=.245\textwidth]{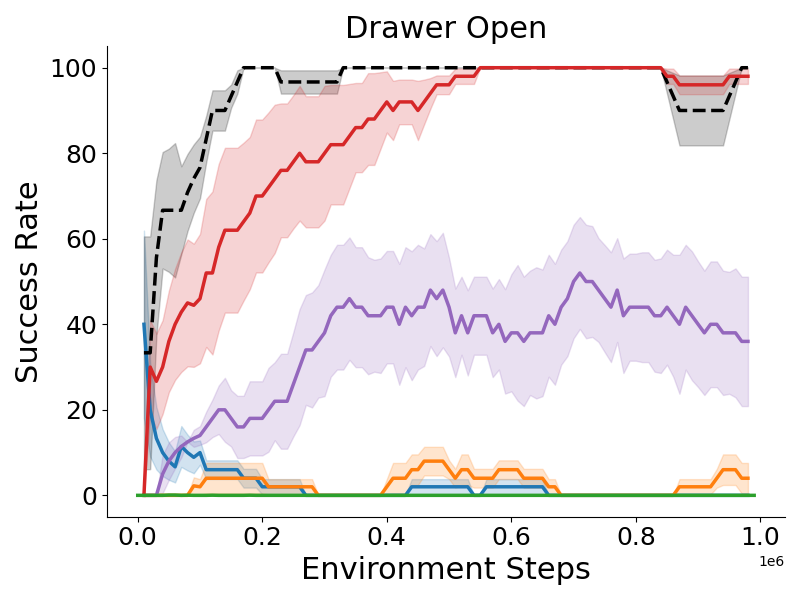}
    \includegraphics[width=.245\textwidth]{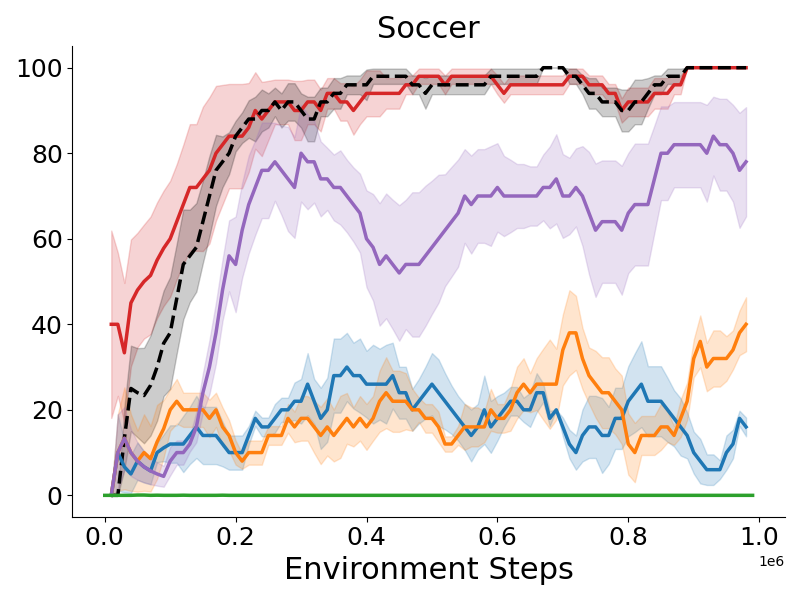}
    \includegraphics[width=.245\textwidth]{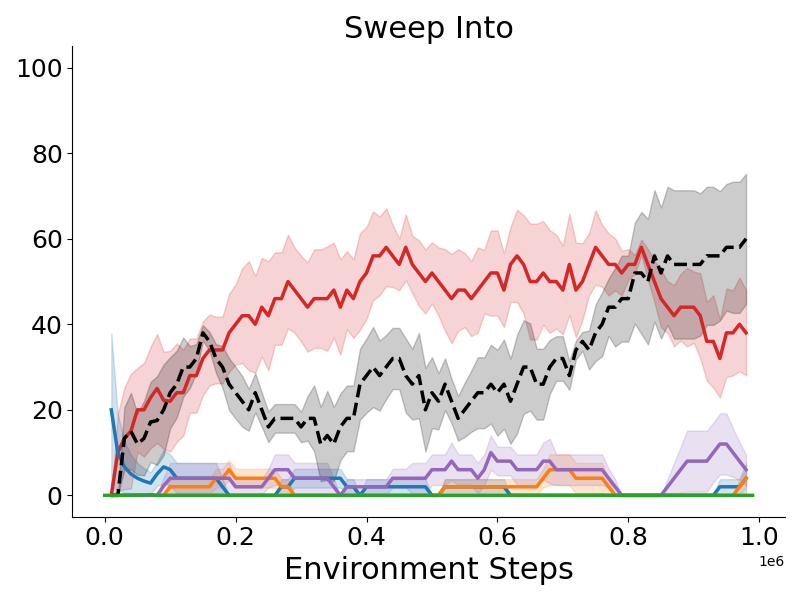} 
    \\
    \includegraphics[width=.245\textwidth]{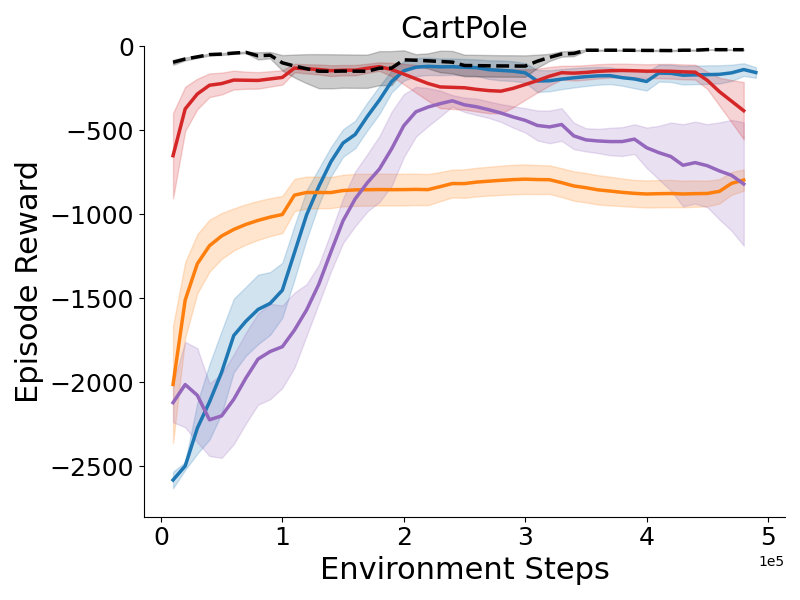}
    \includegraphics[width=.245\textwidth]{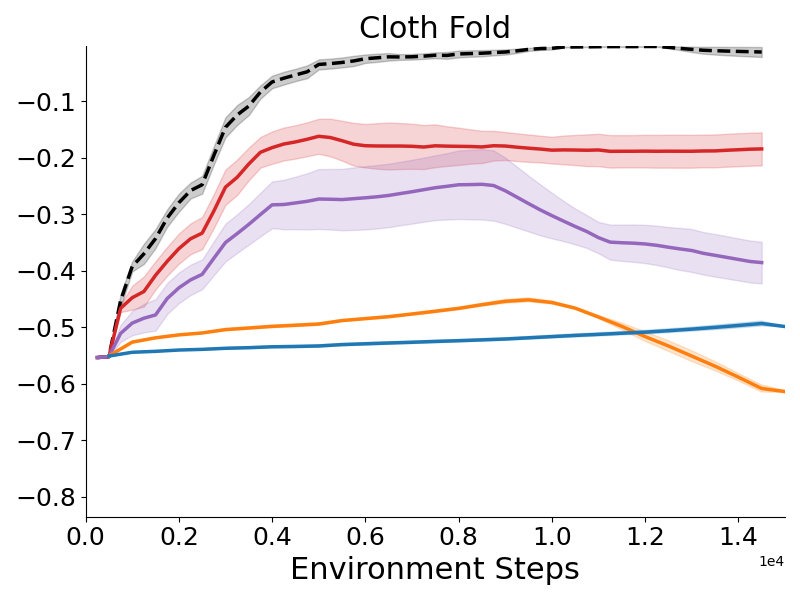}
    \includegraphics[width=.245\textwidth]{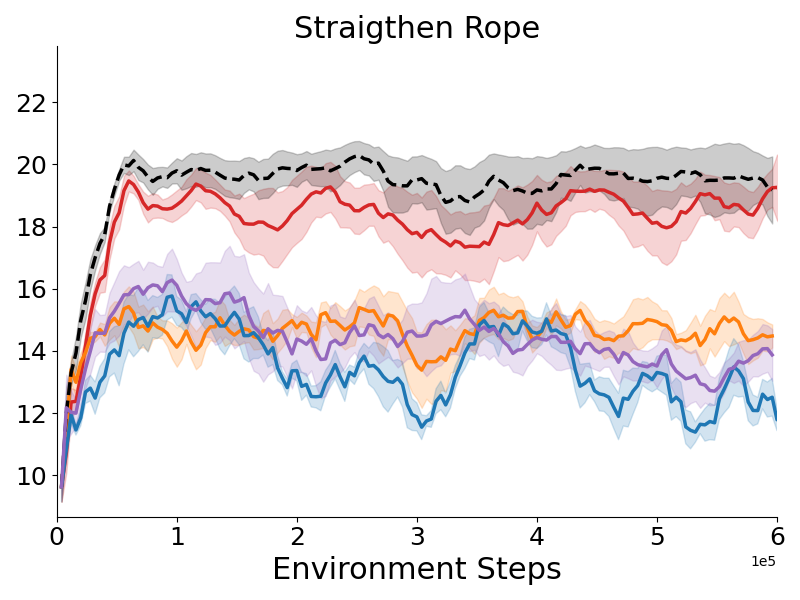}
    \includegraphics[width=.245\textwidth]{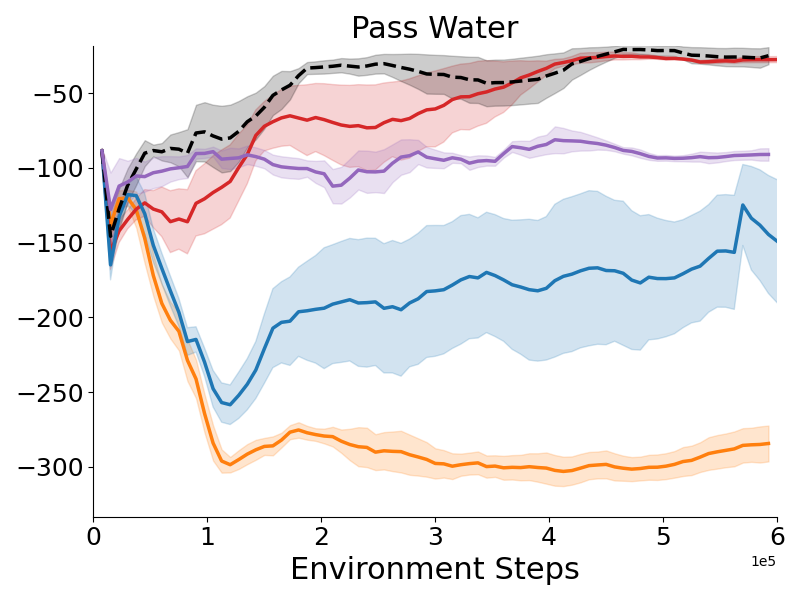}
    \vspace{-0.1in}
    \caption{Learning curves of all compared methods on 7 tasks. \method\ outperforms all baselines in all tasks, and matches or surpasses the performance of GT preference on 6 of the 7 tasks. Results are averaged over 5 seeds, and shaded regions represent standard error. RoboCLIP is only evaluated on the MetaWorld tasks, as this is the set of tasks where the original method is evaluated. }
    \label{fig:learning-curve}
    \vspace{-0.1in}
\end{figure*}

\begin{figure*}[t]
    \centering
    \includegraphics[width=.85\textwidth]{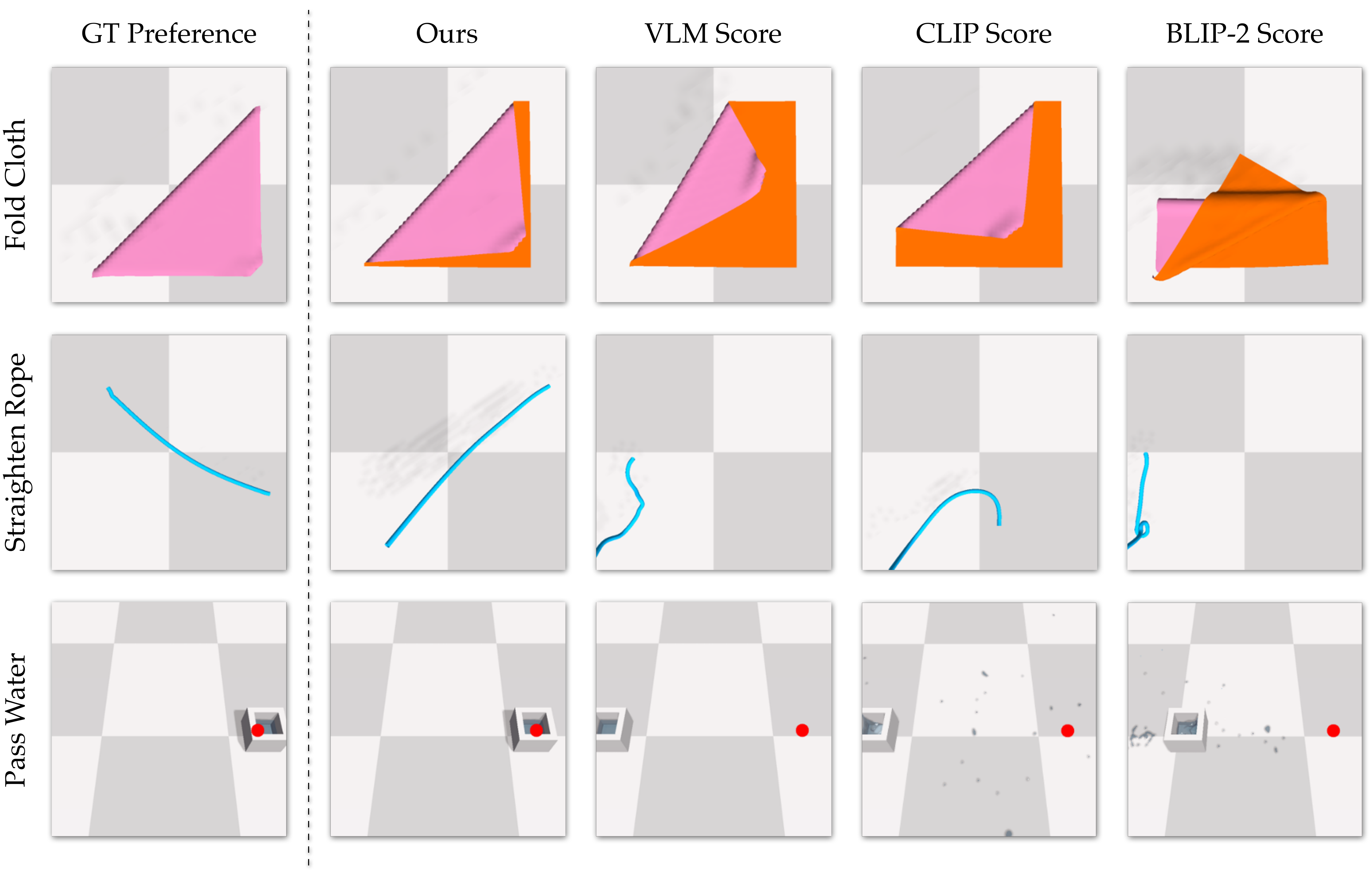}
    \vspace{-0.15in}
    \caption{Comparison of the achieved final state of different methods on SoftGym deformable object manipuation tasks: \textit{Fold Cloth} (Top), \textit{Straighten Rope} (Middle), and \textit{Pass Water} (Bottom). \method\ achieves better final states compared to all the baselines. }
    \label{fig:final-state}
\vspace{-0.15in}
\end{figure*}

\begin{figure*}[t]
    \centering
    \includegraphics[width=.4\textwidth]{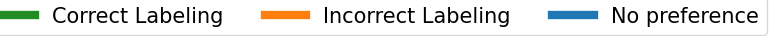} \\     \includegraphics[width=.245\textwidth]{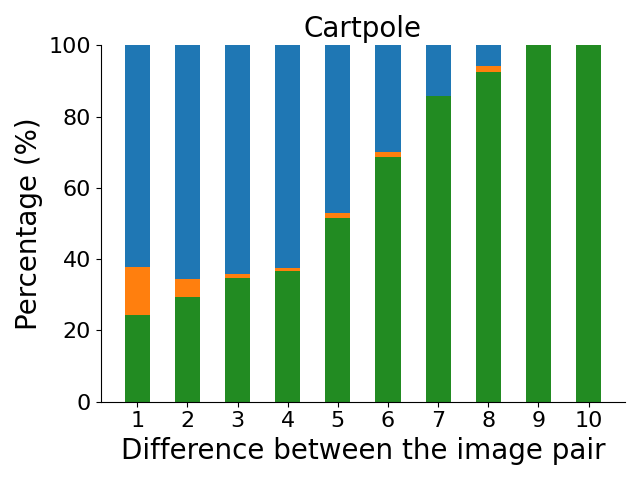}
    \includegraphics[width=.245\textwidth]{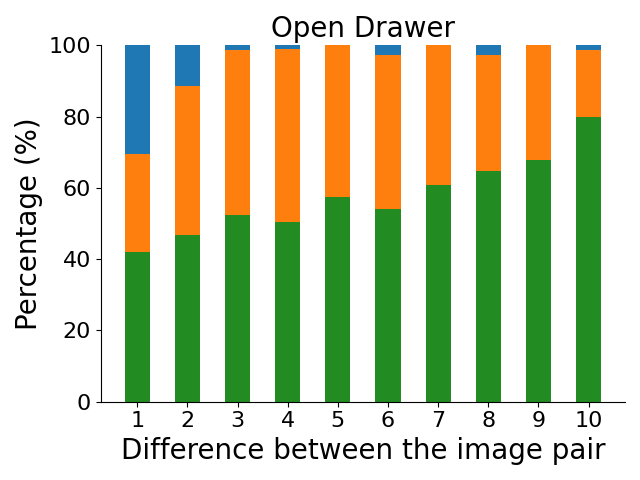}
    \includegraphics[width=.245\textwidth]{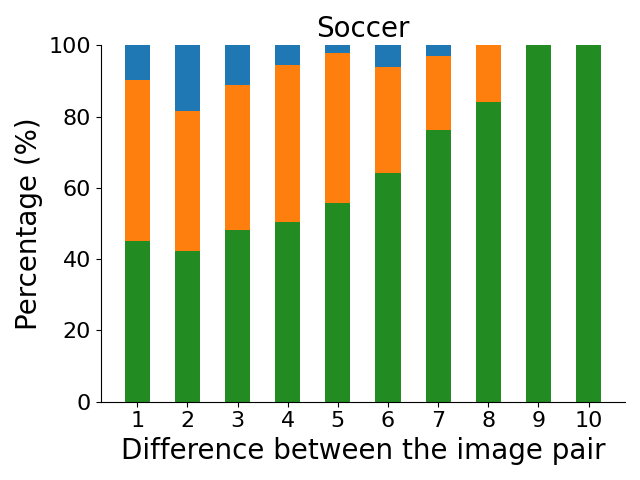}
    \includegraphics[width=.245\textwidth]{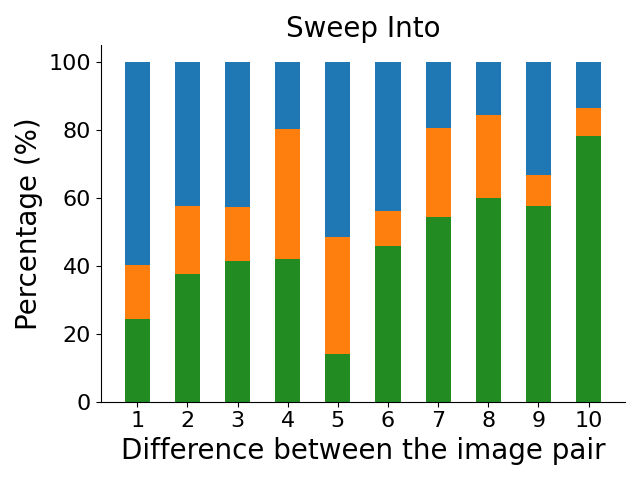}
    \\
    \includegraphics[width=.245\textwidth]{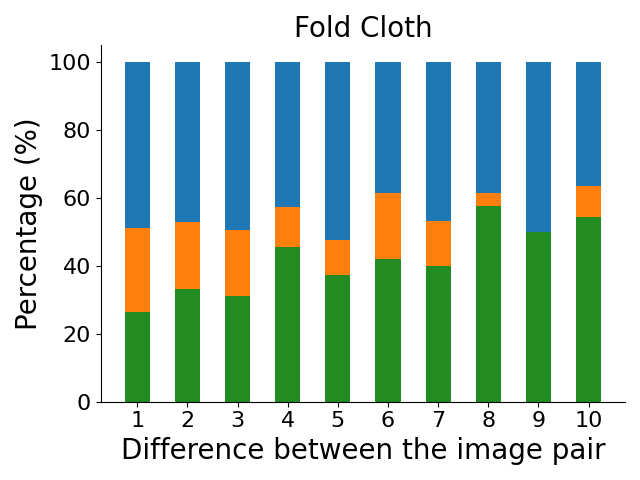}
    \includegraphics[width=.245\textwidth]{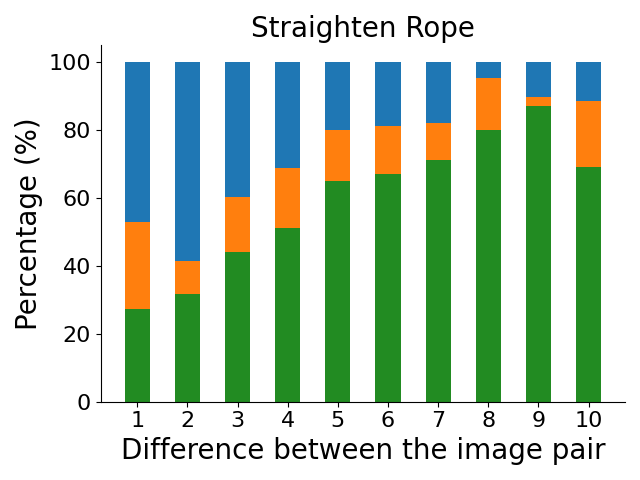}
    \includegraphics[width=.245\textwidth]{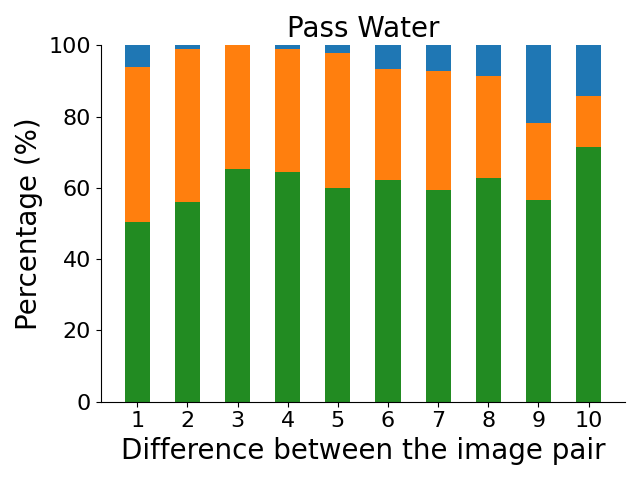}
    \vspace{-0.1in}
    \caption{We provide analysis of the accuracy of the VLM preference labels, compared to ground-truth preference labels defined according to the environment's reward function. The x-axis represents different levels of differences between the image pairs, discretized into 10 bins, where the difference is measured as the difference between the ground-truth task progress associated with the image pairs. The y-axis shows the percentage where the VLM preference labels are correct, incorrect, or when it does not have a preference over the image pairs. 
}
    \label{fig:vlm-acc}
\end{figure*}



\begin{figure*}[t]
    \centering
    \includegraphics[width=.3\textwidth]{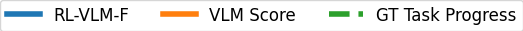} \\
    \includegraphics[width=.3\textwidth]{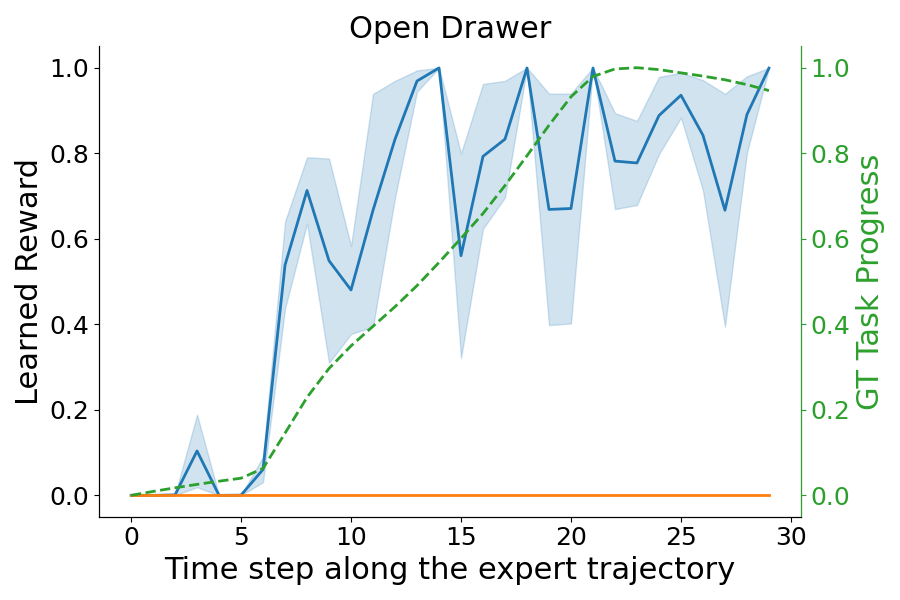}
    ~~~~~
    \includegraphics[width=.3\textwidth]{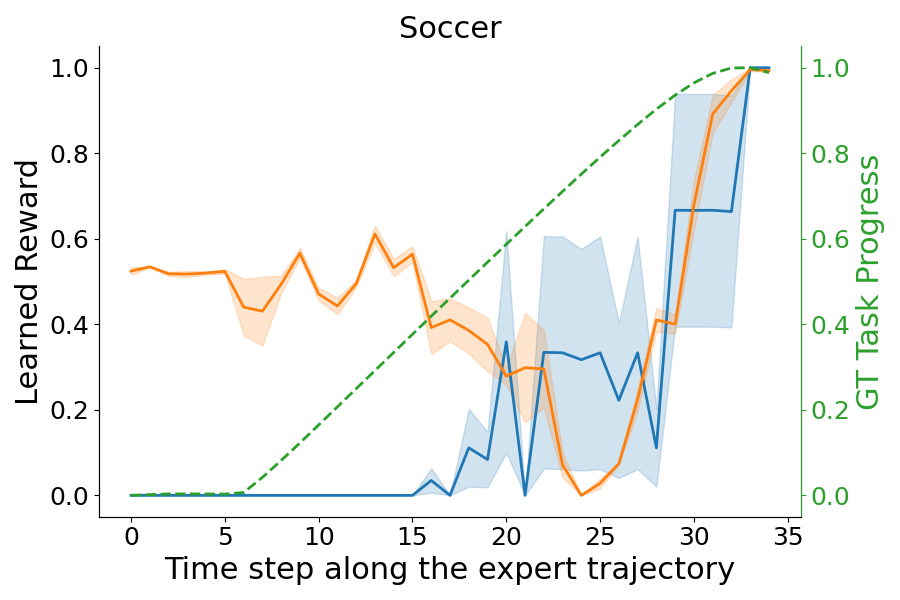}
    ~~~~~
    \includegraphics[width=.3\textwidth]{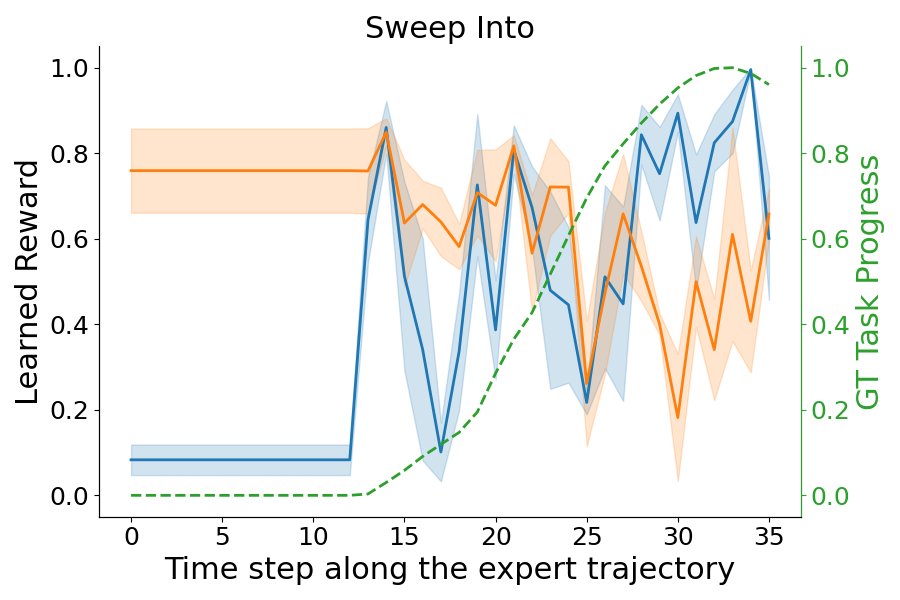}
    \vspace{-0.05in}
    \caption{We compare how well the learned reward by \method\ and VLM Score align with the ground-truth task progress on 3 MetaWorld tasks along an expert trajectory. As shown, \method\ generates rewards that align better with the ground-truth task progress. The learned rewards are averaged over 3 trained reward models with different seeds, and the shaded region represents the standard error. }
\vspace{-0.1in}
    
    \label{fig:learned_reward}
\end{figure*}

\begin{figure*}[t]
    \centering
    \includegraphics[width=.4\textwidth]{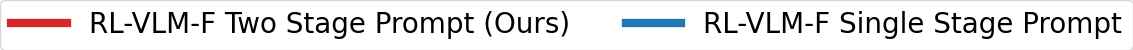} \\
    \includegraphics[width=.24\textwidth]{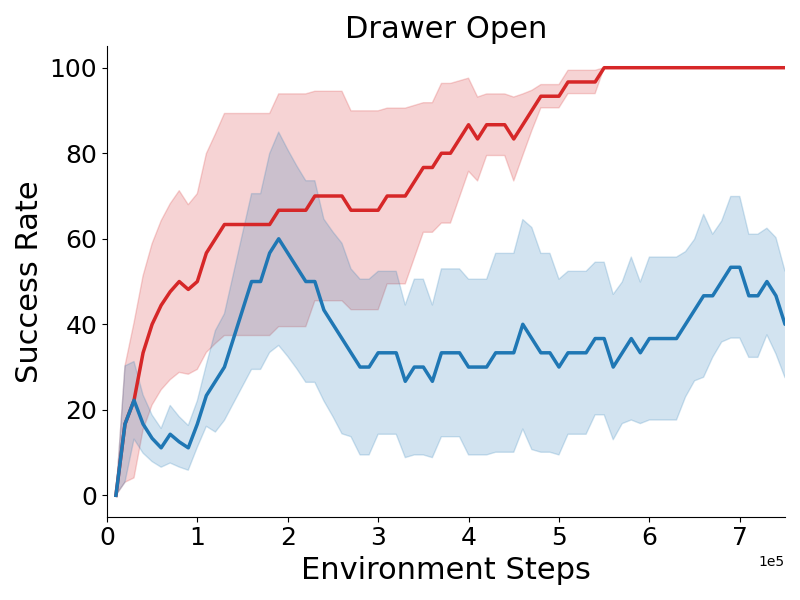}
    \includegraphics[width=.24\textwidth]{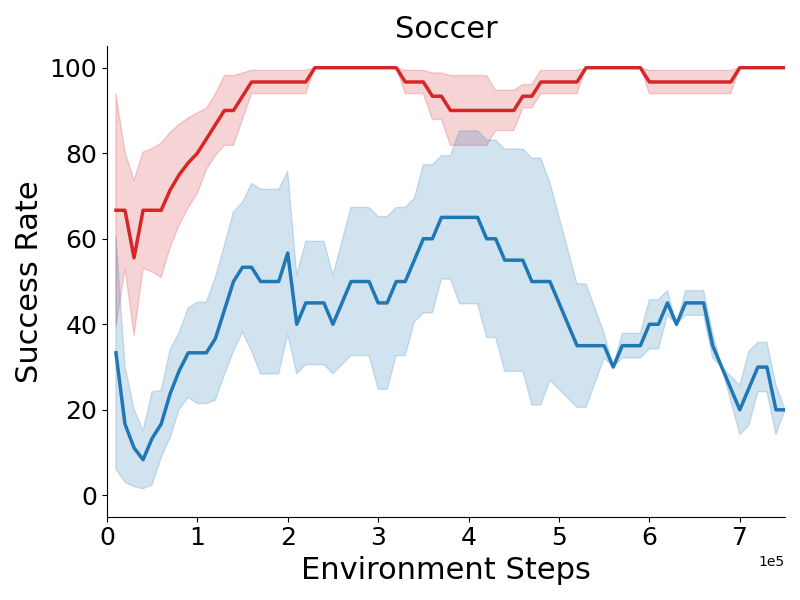}
    \includegraphics[width=.24\textwidth]{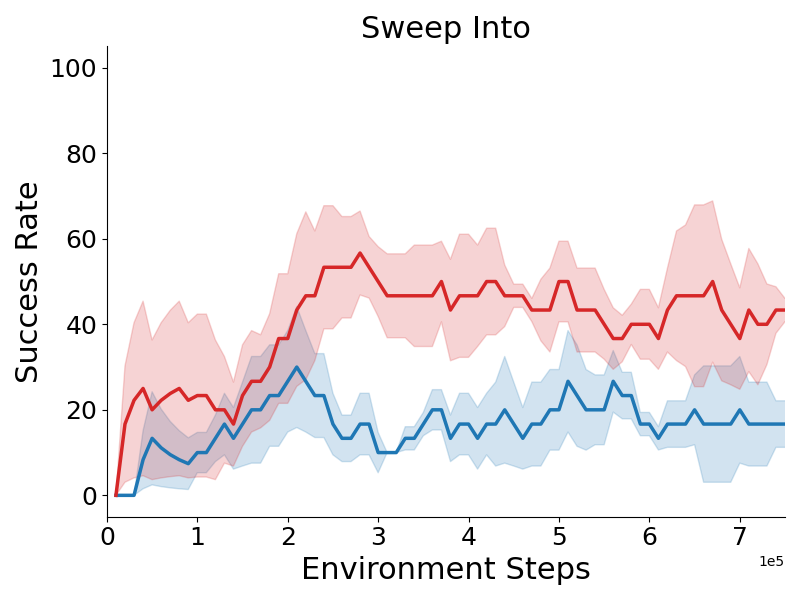}
    \includegraphics[width=.24\textwidth]{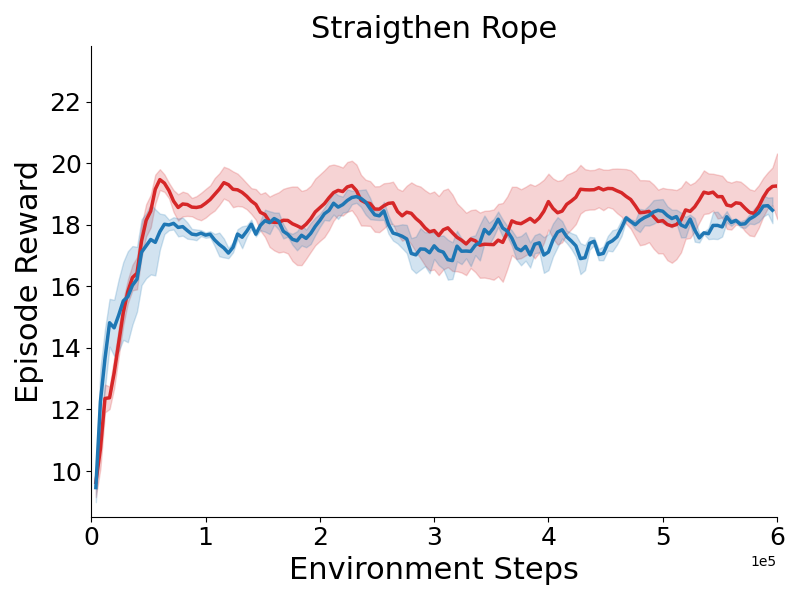}
    \vspace{-0.05in}
    \caption{We compare RL-VLM-F using the proposed two-stage prompting strategy, and an ablated version of using a single-stage prompting strategy. The performance of the single-stage prompting is lower on 3 of the 4 tasks. }
\vspace{-0.1in}
    \label{fig:prompt-ablation}
\end{figure*}



\subsection{Setup}
We evaluate \method\ on a set of tasks, spanning from straightforward classic control tasks to complex manipulation tasks involving rigid, articulated, and deformable objects. 
The tasks are as follows.
\begin{itemize}[nosep]
\item One task from OpenAI Gym~\cite{brockman2016openai}:
\begin{itemize}[nosep]
\item \textit{CartPole} where the goal is to balance a pole on a moving cart.
\end{itemize}
\item Three rigid and articulated object manipulation tasks from MetaWorld~\cite{yu2020meta} with a simulated Sawyer robot:
\begin{itemize}[nosep]
    \item \textit{Open Drawer}, where the robot needs to pull out a drawer;
    \item \textit{Soccer}, where the robot needs to push a soccer ball into a goal; and \item \textit{Sweep Into}, where the robot needs to sweep a green cube into a hole on the table.
\end{itemize}
\item Three deformable object manipulation tasks from SoftGym~\cite{lin2021softgym}:
\begin{itemize}[nosep]
    \item \textit{Fold Cloth}, where the goal is to diagonally fold a cloth from the top left corner to the bottom right corner;
    \item \textit{Straighten Rope}, where the goal is to straighten a rope from a random configuration; and
    \item \textit{Pass Water}, where the goal is to pass a glass of water to a target location without water being spilled out.
\end{itemize}
\end{itemize}
See \Cref{fig:task} for visualizations of these tasks. Further details about the tasks can be found in Appendix~\ref{app:task}. 

We compare to the following baselines that make similar assumptions to us when generating the reward function, i.e., those requiring only a text description and image observations from the agents (without access to environment code). Below is a brief description of each baseline:
\begin{itemize}[nosep]
    \item \textbf{VLM Score}. Instead of querying the VLM to give preference labels over two images, this baseline directly asks the VLM to give a raw score between 0 to 1 for a given image based on the task goal description. We inform the VLM in the prompt that the score should be 1 if the task goal is perfectly achieved in the image. A reward model is then learned to regress to the scores given by the VLM. 
    \item \textbf{CLIP Score} \citep{rocamonde2023visionlanguage}. Given an image, the reward is computed as the cosine similarity score between the embedding of the image and the text description of the task goal using the CLIP model~\cite{radford2021learning}. Such a reward computation method has also been explored in several other prior works~\cite{pmlr-v168-cui22a, mahmoudieh2022zero, adeniji2023language}. 
    \item \textbf{BLIP-2 Score}. Similar to the \textbf{CLIP Score} baseline but uses BLIP-2~\cite{li2023blip} instead of CLIP to compute the cosine similarity score. 
    \item \textbf{RoboCLIP}~\cite{sontakke2023roboclip}. 
    This baseline uses a pre-trained video-language model, S3D~\cite{xie2018rethinking}, to compute the reward as the similarity score between the embedding of the video of the policy trajectories and a demonstration video. 
    Since we do not assume to have access to demonstrations of the task in our method, we use the text version of RoboCLIP for a fair comparison. RoboCLIP-Text uses the pre-trained video-language model to generate rewards as the similarity score between the video embedding of the trajectory and the text embedding of the task description. 
    
    \item \textbf{GT Preference}. We use the original ground-truth reward function (provided by the authors of each benchmark) to give the preference label. This should in theory serve as an oracle and upper bound on the learning performance. 
\end{itemize}
Further details on the baselines, including all the text prompts we use, can be found in Appendices~\ref{app:baseline} and~\ref{app:prompt}. 

For MetaWorld tasks, we use the author-defined task success rate of the policy as the evaluation metric~\cite{yu2020meta}. For all other tasks, we report the episode return of the learned policy. 
For all methods, the policy is learned with state observations, and we use the same policy learning hyper-parameters for all methods, i.e., the only difference between all compared methods is the reward function. 
For methods where a reward function needs to be learned (\method\ and VLM Score), the reward function is learned using image observations. 
For \method\ and the VLM Score baseline, we use Gemini-Pro~\cite{gemini} as the VLM for all tasks except \textit{Fold Cloth}. We find Gemini-Pro to perform poorly on \textit{Fold Cloth}, so we instead use GPT-4V~\cite{gpt4v} as the VLM for this task for these two methods (see Appendix~\ref{app:vlm-quality} for a comparison of Gemini-Pro and GPT-4V on this specific task). We did not run GPT-4V on all tasks due to its quota limitations.  
For all methods except RoboCLIP, we remove the robot from the image for the MetaWorld tasks, as these tasks are all object-centric and removing the robot allows the VLM to focus on the target object when analyzing the images. Since these tasks are simulated, we conveniently use the simulator to make the robot transparent when rendering the images. For real-world applications, techniques such as inpainting can be used to remove the robot from image observations as done in prior work~\cite{bahl2022human, bharadhwaj2023towards}. We keep the robot within the image for RoboCLIP following the original paper's setup. 
We test RoboCLIP only on the MetaWorld tasks, as this is the set of tasks where the original method is evaluated. 

    \vspace{-0.05in}
\subsection{Does \method\ Learn Effective Rewards and Policies?}
We first examine if \method\ leads to useful rewards and policies that can solve the tasks. 
The learning curves of all compared methods on all tasks are shown in Figure~\ref{fig:learning-curve}.
As shown, \method\ outperforms all other baselines in all tasks. 
We find that prior approaches using CLIP or BLIP-2 score can only solve the easiest task -- \textit{CartPole}, and struggle for more complex environments, such as the rigid object manipulation tasks in MetaWorld and the deformable object manipulation tasks in SoftGym. 
The text version of RoboCLIP performs poorly on all three MetaWorld tasks, aligning with the original paper's results, as RoboCLIP works the best with video demonstrations available.  
\method\ also outperforms VLM Score in all tasks, which indicates that prompting VLMs to output a preference label for reward learning results in better task performance in contrast to treating the VLM as a reward function that outputs raw reward scores.  
We also observe that \method\ is able to match the performance of using GT preference in all tasks except Cloth Fold, which suggests we can use a single text description with \method\ to mitigate human efforts in writing complex reward functions for these tasks. 

Interestingly, for the task of \textit{Sweep Into}, the performance of \method\ actually surpasses that of using GT preference. 
We suspect the reason could be as follows: the ground-truth reward function written by the authors for this task includes terms that are not directly correlated to task success. This includes a reward term for grasping the cube, which is not critical for pushing the cube into the hole.
In contrary, \method\ simply uses a text description of the task goal as ``minimize the distance between the cube and the hole'', thus the learned reward is less prone to bias in human-written reward functions and may better reflect the true task goal, leading to better performance.

We show the final states achieved by the policies learned with different methods on the three SoftGym deformable object manipulation tasks in Figure~\ref{fig:final-state}. As shown, for all three tasks, \method\ achieves a final state that is quantifiably better than the baselines.
For \textit{Fold Cloth}, \method\ is closest to a diagonal fold. For \textit{Straighten Rope}, \method\ is able to fully straighten the rope and match the performance of GT preference, where all other baselines failed to fully straighten it. For \textit{Pass Water}, \method\ is able to transport the water to the target location without any water being spilled, and the baselines either do not move the glass, or move it in a way that spills large amounts of water.


\subsection{What is the Accuracy of VLM Preference Labeling?}
Given that \method\ can learn effective rewards and policies that solve the tasks, we perform further analysis on the accuracy of the preference labels generated by a VLM.
To compute accuracy, the VLM outputs $\{-1, 0, 1\}$ (no preference, first image preferred, second image preferred) which we compare to a ground truth preference label defined according to the environment's reward function. 
Note that we discard the image pairs with a label -1 (no preference) when training the reward model. 

Our intuition is that, like humans, it would be hard for the VLM to give correct preference labels when comparing two similar images, and easier to produce correct preference labels when the two images are noticeably dissimilar in terms of achieving the goal. 
Figure~\ref{fig:vlm-acc} presents the accuracy of the VLM at various levels of differences between the two images. 
The ``difference'' between two images is measured as the difference between the ground-truth task progress associated with the images. 
We discretize the differences into 10 bins along the x axis in Figure~\ref{fig:vlm-acc}, where a larger number indicates a greater difference between two images in terms of task progress. 
On the y axis, the green, orange, and blue bars represent the percentage where the VLM preference label is correct, incorrect, or when there is no preference. 
For all tasks, we observe a general trend of increasing accuracy, decreasing uncertainty, and decreasing error as the differences between the images increase, which aligns with intuition.
This trend is most clear and consistent for the \textit{CartPole}, \textit{Open Drawer} and \textit{Soccer} tasks.
Overall, for all tasks, we find that the VLM is able to generate more correct preference labels than incorrect ones, and as shown in Figure~\ref{fig:learning-curve}, the accuracy of VLM-generated preference labels is sufficient for learning a good reward function and policy. 

\subsection{How Does the Learned Reward Align With the Task Progress?}

Figure~\ref{fig:learned_reward} plots the learned rewards (averaged over 3 trained reward models with different random seeds) as well as the true task progress on three MetaWorld environments along an expert trajectory that fully solves the task.
Note the ground-truth task progress is not the same as the author-provided reward function: the author provided reward is a shaped version of the task progress. 
For \textit{Open Drawer}, the task progress is measured as the distance the drawer has been pulled out;
For \textit{Soccer}, it is measured as the negative distance between the soccer ball and the goal;
For \textit{Sweep Into}, it is measured as the negative distance between the cube and the hole.  
We normalize both the ground-truth task progress and the learned reward into the range of [0, 1] for a better comparison between them.
An ideal learned reward should increase as the time step increases along the expert trajectory, as like the ground-truth task progress. 
As shown, the reward learned by \method\ aligns better with the ground-truth task progress compared with the VLM Score baseline. 
We do observe that the learned reward tends to be noisy and includes many local minima. Despite this, the learned reward still achieves the highest value when the task progresses the most (except for the task of Sweep Into). As shown in Figure~\ref{fig:learning-curve}, the learned reward is sufficient for learning successful policies.
For \textit{Open drawer}, we notice that the reward produced by VLM Score remains zero. This is likely because, during training, most of the scores given by the VLM are 0, and the model learns to predict 0 at all time steps to minimize the regression loss.
We find the CLIP and BLIP-2 scores on these environments are generally noisy; the corresponding plots can be found in Appendix~\ref{app:learned_reward}.  

\subsection{Ablation on the Prompt Strategy}
\label{sec:prompt-ablation}
We used a two-stage prompting strategy for \method{}, where the VLM is first asked to analyze the pair of images in the analysis stage, and then output the preference label in the labeling stage. 
Here we compare it with a single-stage prompting strategy where we query the VLM to directly output a preference label over the two image observations in a single stage. 
The detailed single-stage prompt can be found in Appendix~\ref{app:prompt-ablation}. Figure~\ref{fig:prompt-ablation} presents the comparison on 4 tasks: Open Drawer, Soccer, Sweep Into and Straighten Rope. As shown, the success rate of using the VLM with the single-stage prompt is lower than that of using the two-stage prompt on 3 out of the 4 tasks.






\section{Conclusion and Future Work}
In this work, we present \method, a method that automatically generates reward functions via querying VLMs with preferences given a task descriptions and image observations for a wide range of tasks. We demonstrate our proposed method's effectiveness on rigid, articulated, and deformable object manipulation tasks. 

Future work could extend \method\ to an active learning context, exploring both easy and informative VLM queries for more efficient reward learning. The adaptable nature of our method allows for the integration of more advanced VLMs when they become available, potentially addressing more complex tasks. It would also be interesting to test \method\ on tasks with a longer horizon. One could first decompose the tasks into subtasks with shorter horizons, either via manual decomposition or foundation models~\cite{ahn2022can}. Then, \method\ can be used to solve each subtask. Additionally, our approach offers a practical pathway to applying RL in real-world settings, where obtaining reward functions is often difficult.

\section*{Acknowledgements}
This work is supported by the National Science Foundation under Grant No. IIS-2046491. Any opinions, findings, and
 conclusions or recommendations expressed in this material are
 those of the author(s) and do not necessarily reflect the views
 of the National Science Foundation. 

\section*{Impact Statement}
As we use pre-trained Vision Language Models for generating the reward functions, the bias presented in the VLMs might be inherited into the reward function and subsequently the learned policy. As a result, one might want to examine the behavior of the learned policy before deploying it to safety critical applications. Other than this point, we do not anticipate any societal consequences of our work that must be specifically highlighted here.

\balance
\bibliography{ref}
\clearpage
\newpage
\appendix
\section*{Appendix} 
\addcontentsline{toc}{section}{Appendix} 

\definecolor{lightgray}{gray}{0.9}
\definecolor{darkgray}{gray}{0.8}

\section{Details on Tasks and Environments} 
\label{app:task}
We run our method and baselines on \textit{\textit{CartPole}} from openAI Gym~\cite{brockman2016openai},
three rigid and articulated object manipulation tasks from MetaWorld~\cite{yu2020meta}, and three deformable object manipulation tasks from SoftGym~\cite{lin2021softgym}. 
For the three MetaWorld tasks, we modified the gripper initial state such that it starts close to the target object to manipulate. Figure 3 in the paper shows the initial state for these 3 tasks. We also adjusted the camera view such that the target object is clearly visible at around the center of the image, to provide good images for VLM to give preferences.
We describe the observation space and action space for those tasks as follows:
\renewcommand{\thesubsection}{\thesection.\arabic{subsection}}


\subsection{Observation Space} 
For policy learning with SAC, we use state-based observations; for reward learning, we use high dimensional RGB image observations, rendered by the simulator. We now detail the state-based observation space for each task.

\textbf{MetaWorld Tasks.}
For MetaWorld tasks, we follow the setting in the original paper~\cite{yu2020meta}. The state observation always has 39 dimensions. It consists of the position and gripper status of the robot's end-effector, the position and orientation of objects in the scene, and the position of the goal.  

\textbf{\textit{CartPole.}}
The state observation has 4 dimensions, including the position and velocity of the cart, as well as the angle and angular velocity of the pole.

\textbf{\textit{Cloth Fold.}}
The state observation is the position of a subset of the  particles in the cloth mesh. The cloth is of size 40 x 40, and we uniformly subsample it to be of size 8 x 8. The state is then the position of the picker, and the positions of all those subsampled particles.

\textbf{\textit{Straighten Rope.}}
The state observation is the positions of all particles on the rope and has 36 dimensions.

\textbf{\textit{Pass Water.}}
The state observation includes the size (width, length, height) of the container, the target container position, height of the water in the container, amount of water inside and outside of the container. The state observation has 7 dimensions.

\subsection{Action Space} 
For all environments, we normalize the action space to be within $[-1,1]$. Below we describe the action space for each environment.

\textbf{MetaWorld Tasks.}
For MetaWorld tasks, the action space always has four dimensions. It includes the change in 3D position of the robot's end-effector followed by a normalized torque that the gripper fingers should apply.

\textbf{\textit{CartPole.}}
The original action space is a discrete value in ${0,1}$, indicating the direction of the fixed force the cart is pushed with. We modified it to be continuous within range [0, 1] such that SAC can be used as the learning algorithm. The continuous action represents the force applied to the pole. 

\textbf{\textit{Cloth Fold.}}
For this task, we use a pick-and-place action primitive. We assume that the corner of the cloth is grasped when the task is initialized. The action is the 2D target place location. 

\textbf{\textit{Straighten Rope.}}
For this task, we use two pickers, one at each end of the rope, to control the rope. Therefore, the action space is the 3D delta positions for each picker and has 6 dimensions in total. We assume the two end points of the rope is already grasped at the beginning of the task. 

\textbf{\textit{Pass Water.}}
The motion of the glass container is constrained to be in one dimension. Therefore, the action also has a dimension of 1 and is the delta position of the container along the dimension.

\section{Hyper-parameters and Network Architectures} 
\label{app:parameter}
\subsection{Image-based Reward Learning}
For the image-based reward model, we use a 4-layer Convolutional Neural Network for MetaWorld tasks and \textit{CartPole} and a standard ResNet-18~\cite{he2016deep} for the three deformable object manipulation tasks. Following PEBBLE~\cite{lee2021pebble}, we also use an ensemble of three reward models and use tanh as the activation function for outputting reward. For \method, we train the model by optimizing the cross-entropy loss, defined in Equation~\ref{eq:reward_function}. For VLM Score, we train the mode by optimizing the MSE loss between the predicted score and ground-truth score output by the VLM. For both methods, we use ADAM~\cite{kingma2014adam} as the optimizer with an initial learning rate of $0.0003$. 

\subsection{Policy Learning}
Following PEBBLE~\cite{lee2021pebble}, we use 
SAC as the off-policy learning algorithm. We follow the network architectures for the actor and critic and all the hyper-parameter settings in the original paper for policy learning.

\subsection{Training details}
Our implementation is based on PEBBLE~\cite{lee2021pebble}. Below we describe the feedback collection schedule for each task. For all tasks, we use a segment size of $1$. We summarize the number of queries per feedback session ($M$ in Algorithm~\ref{algo}), the frequency at which we collect feedback in terms of environment steps ($K$ in Algorithm~\ref{algo}), and the maximum budget of queries ($N$) for each task in Table~\ref{tab:hyperparam_feedback_schedule}. For Cloth Fold, we have to use a small number of maximum budget of queries due to the quota limitation of GPT-4V.

\begin{table}[t]
\centering
\begin{tabular}{|c|c|c|c|}
\hline
  & M & K & N \\ \hline
\textit{Open Drawer} &   40 & 4000  & 20000   \\ \hline
\textit{\textit{Soccer}} &   40 &  4000 & 20000  \\ \hline
\textit{Sweep Into} &  40 & 4000  & 20000  \\ \hline
\textit{CartPole} & 50  & 5000  & 10000   \\ \hline
\textit{Cloth Fold} & 50 & 1000   &  500 \\ \hline
\textit{Straighten Rope} & 100  & 5000  &  12000 \\ \hline
\textit{Pass Water} &  100 & 5000  & 12000  \\ \hline
\end{tabular}
\caption{Hyper-parameters for feedback learning schedule.}
\label{tab:hyperparam_feedback_schedule}
\end{table}

\begin{figure}[t]
    \centering
    \includegraphics[width=.8\columnwidth]{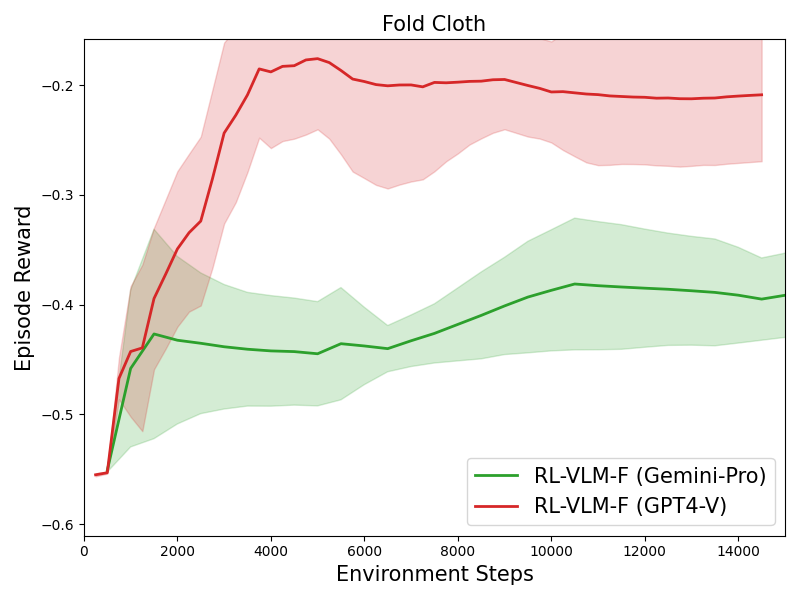}
    \vspace{-0.2in}
    \caption{On the Fold Cloth task, we find the performance of GPT-4V to be better than Gemini-Pro, possibly due to the complex visual appearance of the cloth. }
    \label{fig:vlm-quality}
\end{figure}

\section{Baselines} 
\label{app:baseline}

\subsection{VLM score}
For this baseline, we use the same amount of queries ($K$) at the same frequency ($M$) as in our method to ask VLM to directly output a score between $0$ to $1$. The reward model's architecture is the same as our method, except that the model is trained with regression loss to regress to VLM's output score instead of classification loss as done in our method. 

\subsection{RoboCLIP}
In RoboCLIP, the backbone video-language model is S3D~\cite{xie2018rethinking}, trained on clips of human activities paired with textual descriptions from the HowTo100M dataset~\citep{miech19howto100m}. Given the assumption that the model generalizes to unseen robotic environments, we applied this baseline solely to the three MetaWorld tasks that contain a robot in the scene. We obtain the implementation directly from the authors. To maintain uniform assumptions across methods, we compare against the RoboCLIP variant that only uses a text description instead of a video demonstration to compute the similarity score with the agent's episode rollout for reward computation. According to the original paper, this text-only variant of RoboCLIP underperforms the video-based method, corroborating the lower performance observed in our tasks.

\section{Prompts}
\label{app:prompt}
\subsection{\method\ and VLM Score}
For both \method\ and VLM Score, we use a unified query template combined with specific task goal descriptions. The templates for \method\ and VLM Score are shown in Figure~\ref{fig:Prompt Template for ours} and Figure~\ref{fig:Prompt Template for VLM Score}: 



The only task-specific part in both prompts is the task goal description. We use the same set of descriptions for both methods. We summarize the textual description for each task in Table~\ref{tab:task_goal_description_ours_and_vlm_score}.

\subsection{CLIP Score and BLIP-2 Score}
The task descriptions for both CLIP Score and BLIP-2 Score baselines are summarized in Table~\ref{tab:task_goal_description_clip_and_blip2}. The semantic meaning is almost identical to those used by \method\ and VLM Score, except that the description is structured differently. 
For \textit{CartPole}, we used the exact same prompt as in~\cite{rocamonde2023visionlanguage}, since they reported successful learning of this task using that prompt. 

\subsection{RoboCLIP}
For the task descriptions for the RoboCLIP baseline, we followed the format used in the original paper~\cite{sontakke2023roboclip}. We summarize the text descriptions in Table~\ref{tab:task_goal_description_RoboCLIP}.

\begin{figure*}[t]
    \centering
    \includegraphics[width=0.9\textwidth]{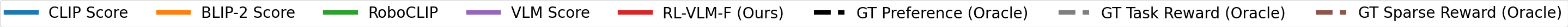}\\
    \includegraphics[width=.245\textwidth]{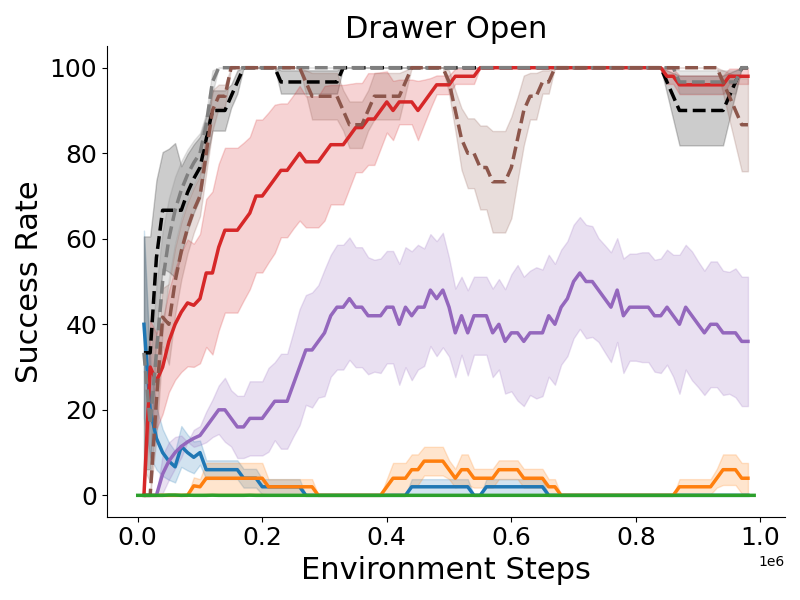}
    \includegraphics[width=.245\textwidth]{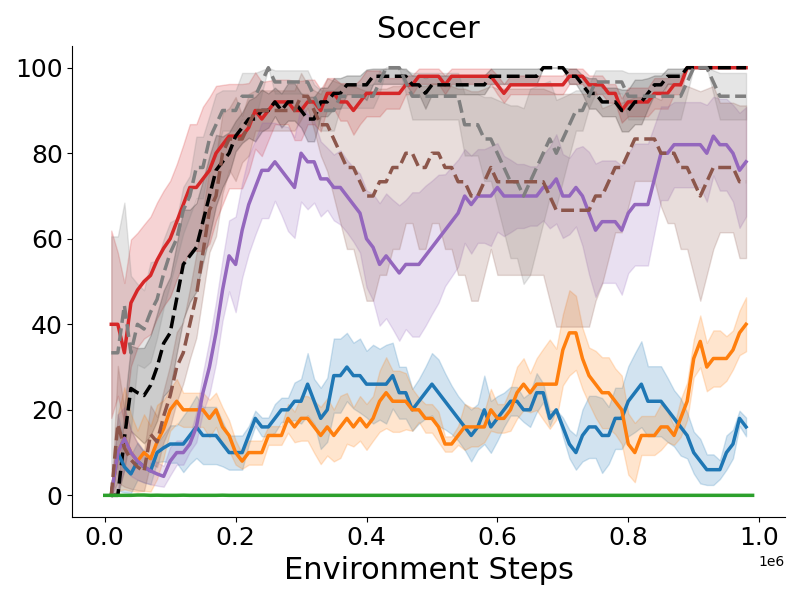}
    \includegraphics[width=.245\textwidth]{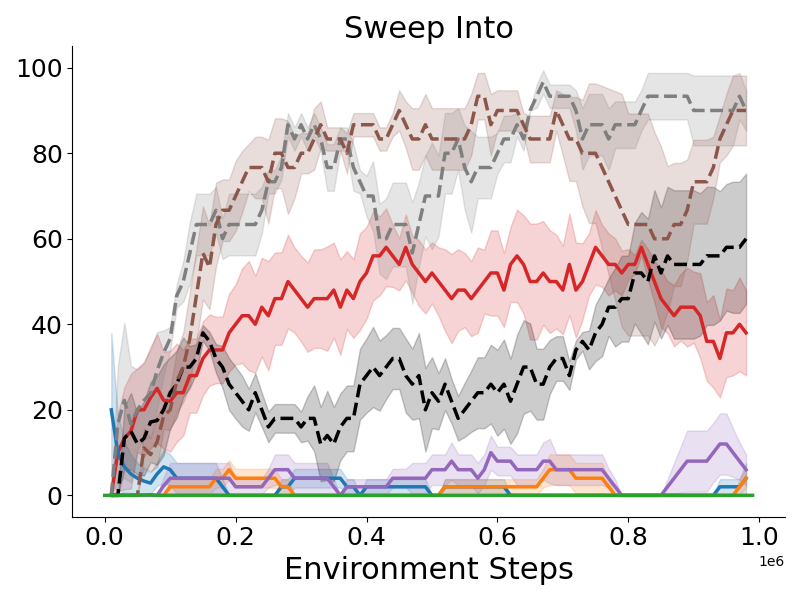} 
    \\
    \includegraphics[width=.245\textwidth]{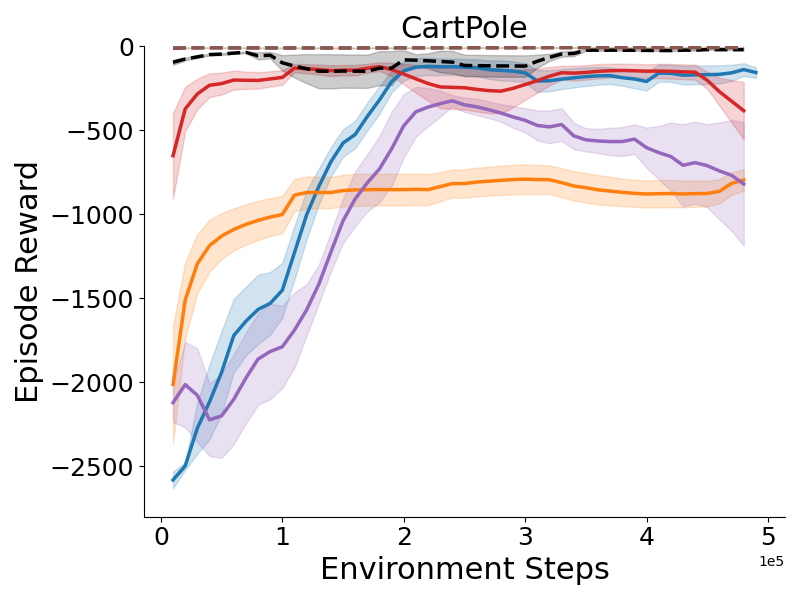}
    \includegraphics[width=.245\textwidth]{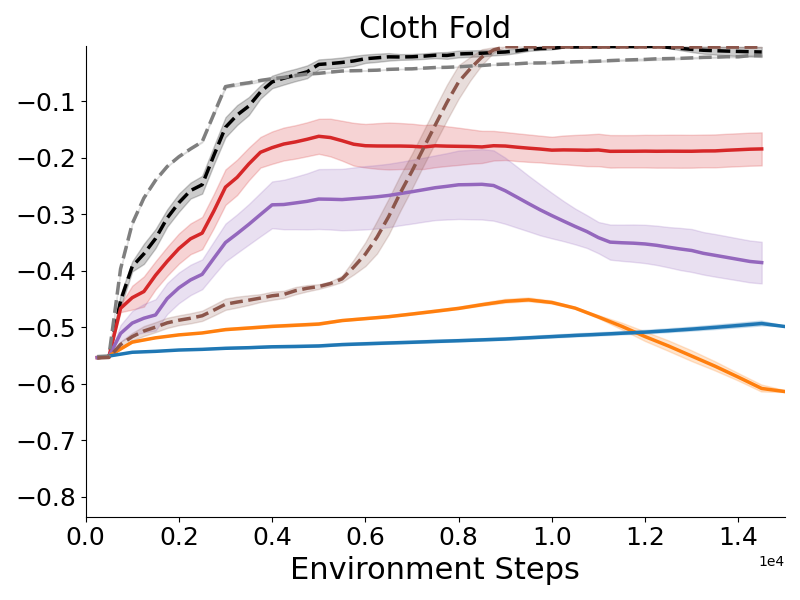}
    \includegraphics[width=.245\textwidth]{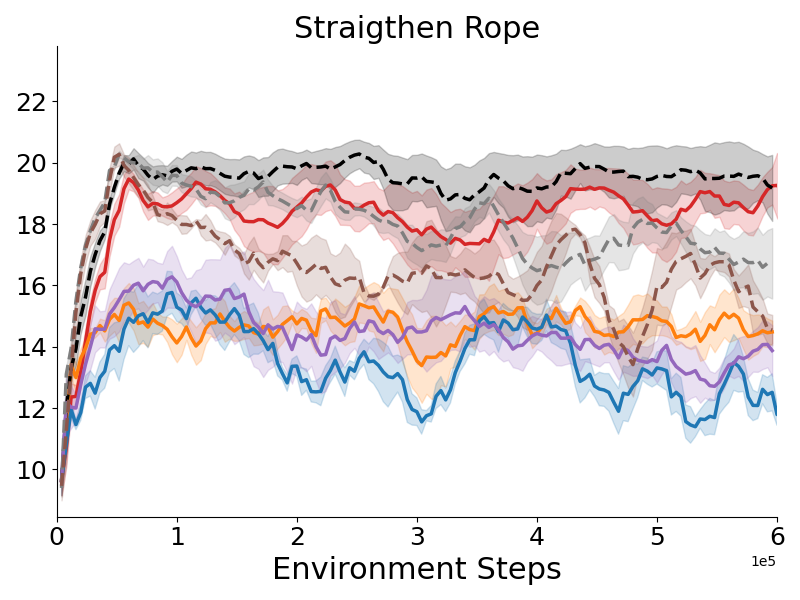}
    \includegraphics[width=.245\textwidth]{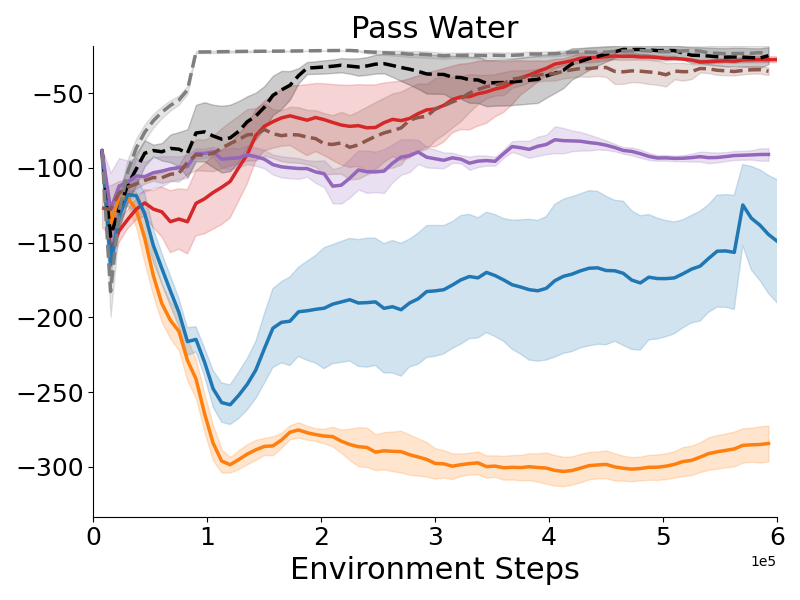}
    \vspace{-0.1in}
    \caption{Learning curves of GT Task Reward (Oracle) and GT Sparse Reward (Oracle), along with \method\ and all baselines. }
    \label{fig:learning-curve-include-dense-sparse}
    \vspace{-0.1in}
\end{figure*}

\subsection{RL-VLM-F single stage prompt}
\label{app:prompt-ablation}
In Section~\ref{sec:prompt-ablation} we compared to an ablated version of RL-VLM-F where a single-stage prompting strategy is used. The single-stage prompt used is shown in Figure~\ref{fig:prompt-single-stage}. For a fair comparison, it is kept to be the same as the two-stage prompt with only minor differences.

\section{Additional Experiment Results}
\label{app:full-experiment-results}

\subsection{GT Task Reward (Oracle) and GT Sparse Reward (Oracle)}
To better contextualize the results from different reward models, we test two more baselines, i.e., GT Task Reward (Oracle) and GT Sparse Reward (Oracle). For GT Task Reward (Oracle), we use the original ground-truth human-written reward function with SAC as the RL algorithm to train the policy. For GT Sparse Reward (Oracle), we use sparse reward with SAC. The reward is 1 when the goal is achieved and 0 otherwise. The results of GT Task Reward (Oracle) and GT Sparse Reward (Oracle), along with our method and all baselines, are shown in Figure~\ref{fig:learning-curve-include-dense-sparse}. For most tasks, \method\ 's final performance can match that of using ground-truth reward, highlighting the effectiveness of our method.

\begin{figure*}[t]
\begin{prompttextbox}[title=Prompt Template for \method\ (ours)]
\textbf{Analysis Template}

Consider the following two images:\\
Image 1:\\
\textcolor{red}{[Image 1]}\\
Image 2:\\
\textcolor{red}{[Image 2]}\\

1. What is shown in Image 1?\\
2. What is shown in Image 2? \\
3. The goal is to \textcolor{red}{[task description]}. Is there any difference between Image 1 and Image 2 in terms of achieving the goal?\\

\textbf{Labeling Template}

Based on the text below to the questions:\\
\textcolor{red}{[Repeat the 3 questions in the Analysis Template]}\\
\textcolor{red}{[VLM response]}\\
Is the goal better achieved in Image 1 or Image 2? Reply a single line of 0 if the goal is better achieved in Image 1, or 1 if it is better achieved in Image 2.\\
Reply -1 if the text is unsure or there is no difference.
\end{prompttextbox}
\caption{Prompt Template for \method.}
\label{fig:Prompt Template for ours}
\end{figure*}


\begin{figure*}[t]
\begin{prompttextbox}[title=Single Stage Prompt Template for \method ]

Consider the following two images:\\
Image 1:\\
\textcolor{red}{[Image 1]}\\
Image 2:\\
\textcolor{red}{[Image 2]}\\

1. What is shown in Image 1?\\
2. What is shown in Image 2?\\
3. The goal is \textcolor{red}{[task description]}. Is there any difference between Image 1 and Image 2 in terms of achieving the goal? \\

Is the goal better achieved in Image 1 or Image 2?
Reply a single line of 0 if the goal is better achieved in Image 1, or 1 if it is better achieved in Image 2.\\
Reply -1 if the text is unsure or there is no difference.
\end{prompttextbox}
\caption{The single stage prompt Template for \method.}
\label{fig:prompt-single-stage}
\end{figure*}

\begin{figure*}[t]
\begin{prompttextbox}[title=Prompt Template for VLM Score]
\textbf{Analysis Template}
Consider the following image:\\
\textcolor{red}{[Image]}\\
1. What is shown in the image?\\
2. The goal is \textcolor{red}{[task description]}. On a scale of 0 to 1, the score is 1 if the goal is achieved. What score would you give the image in terms of achieving the goal?\\
\textbf{Labeling Template}

Based on the text below to the questions:\\
\textcolor{red}{[Repeat the 3 questions in the Analysis Template]}\\
\textcolor{red}{[VLM response]}\\
Please reply a single line of the score the text has given.
Reply -1 if the text is unsure.
\end{prompttextbox}
\caption{Prompt Template for VLM Score.}
\label{fig:Prompt Template for VLM Score}
\end{figure*}

\begin{table*}[t]
  \centering
  \begin{tabularx}{\textwidth}{>{\bfseries}l X}
    \toprule
    \rowcolor{lightgray}
    Task Name & Goal Description \\
    \rowcolor{darkgray}
    \textit{Open Drawer} & to open the drawer \\
    \rowcolor{lightgray}
    \textit{Soccer} & to move the soccer ball into the goal \\
    \rowcolor{darkgray}
    \textit{Sweep Into} & to minimize the distance between the green cube and the hole \\
    \rowcolor{lightgray}
    \textit{CartPole} & to balance the brown pole on the black cart to be upright \\
    \rowcolor{darkgray}
    \textit{Cloth Fold} & to fold the cloth diagonally from top left corner to bottom right corner \\
    \rowcolor{lightgray}
    \textit{Straighten Rope} & to straighten the blue rope \\    
    \rowcolor{darkgray}
    \textit{Pass Water} & to move the container, which holds water, to be as close to the red circle as possible without causing too many water droplets to spill \\  
    \bottomrule
    \end{tabularx}
  \caption{Goal description used in \method\ and VLM Score baseline.}
  \label{tab:task_goal_description_ours_and_vlm_score}
\end{table*}


\begin{table*}[t]
  \centering
  \begin{tabularx}{\textwidth}{>{\bfseries}l X}
    \toprule
    \rowcolor{lightgray}
    Task Name & Goal Description \\
    \rowcolor{darkgray}
    \textit{Open Drawer} & The drawer is opened. \\
    \rowcolor{lightgray}
    \textit{Soccer} & The soccer ball is in the goal. \\
    \rowcolor{darkgray}
    \textit{Sweep Into} & The green cube is in the hole. \\
    \rowcolor{lightgray}
    \textit{CartPole} & pole vertically upright on top of the cart. \\
    \rowcolor{darkgray}
    \textit{Cloth Fold} & The cloth is folded diagonally from top left corner to bottom right corner. \\
    \rowcolor{lightgray}
    \textit{Straighten Rope} & The blue rope is straightened. \\    
    \rowcolor{darkgray}
    \textit{Pass Water} & The container, which holds water, is as close to the red circle as possible without causing too many water droplets to spill. \\  
    \bottomrule
    \end{tabularx}
  \caption{Goal description used in CLIP Score and BLIP-2 Score.}
  \label{tab:task_goal_description_clip_and_blip2}
\end{table*}

\begin{table*}[t]
  \centering
  \begin{tabularx}{\textwidth}{>{\bfseries}l X}
    \toprule
    \rowcolor{lightgray}
    Task Name & Goal Description \\
    \rowcolor{darkgray}
    \textit{Open Drawer} & robot opening green drawer \\
    \rowcolor{lightgray}
    \textit{Soccer} & robot pushing the soccer ball into the goal \\
    \rowcolor{darkgray}
    \textit{Sweep Into} & robot sweeping the green cube into the hole on the table \\
    \bottomrule
    \end{tabularx}
  \caption{Goal description used in RoboCLIP.}
  \label{tab:task_goal_description_RoboCLIP}
\end{table*}

\subsection{Ablation Study: Influence of Using Different VLMs} 
\label{app:vlm-quality}
For \method\ and the VLM score baseline, we use Gemini-Pro~\cite{gemini} as the VLM for all tasks except Fold Cloth. We find Gemini-Pro to perform poorly on Fold Cloth, so we instead use GPT-4V~\cite{gpt4v} as the VLM for this task for both methods. Figure~\ref{fig:vlm-quality} compares the learning performance of Gemini-Pro versus GPT-4V on the task of Fold Cloth. We do observe GPT-4V to achieve much better performance on this task than Gemini-Pro. The poorer performance of Gemini-Pro on this task could be possibly due to the more complex visual reasoning required for deformable cloth.


\subsection{More Visualization of the Learned Reward}
\label{app:learned_reward}
\begin{figure*}[t]
    \centering
    \includegraphics[width=.245\textwidth]{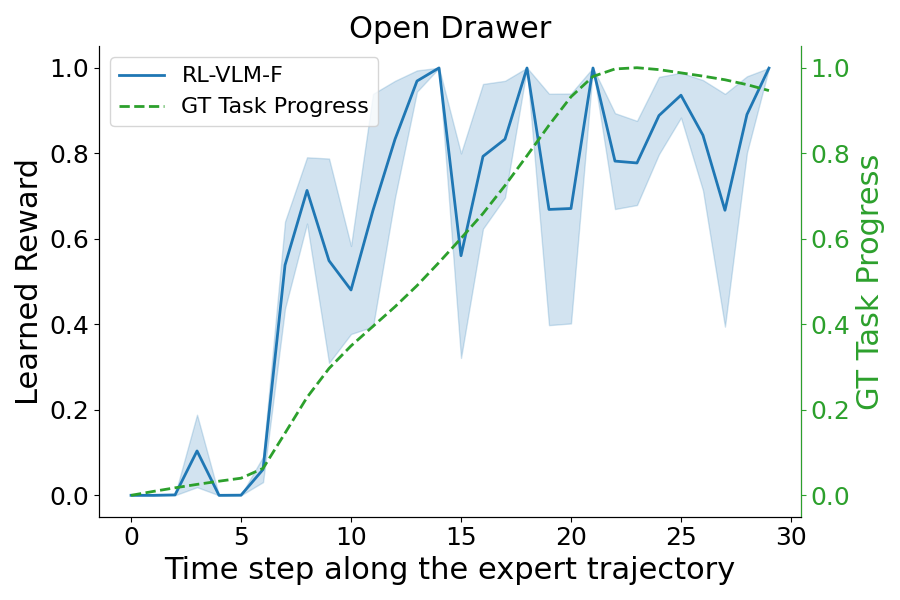}
    \includegraphics[width=.245\textwidth]{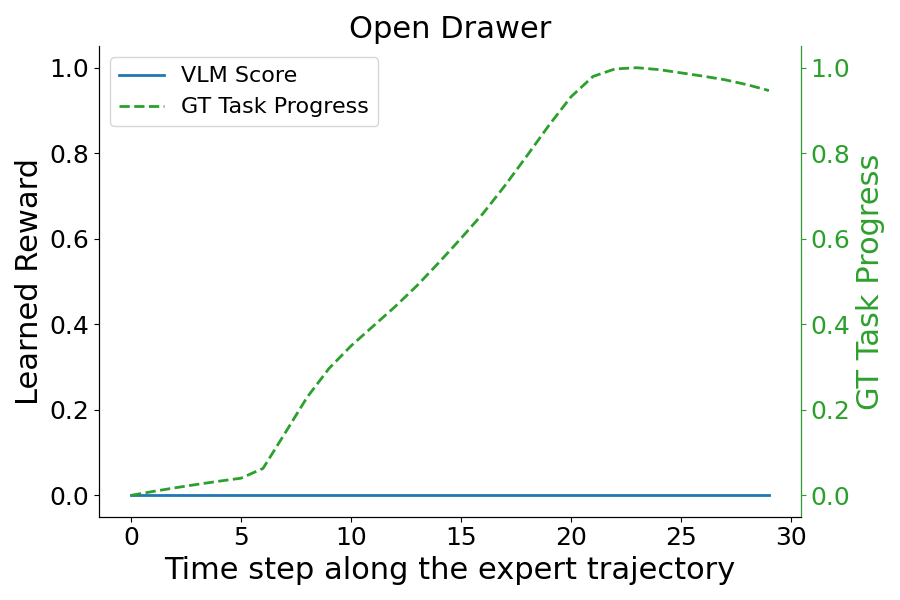}
    \includegraphics[width=.245\textwidth]{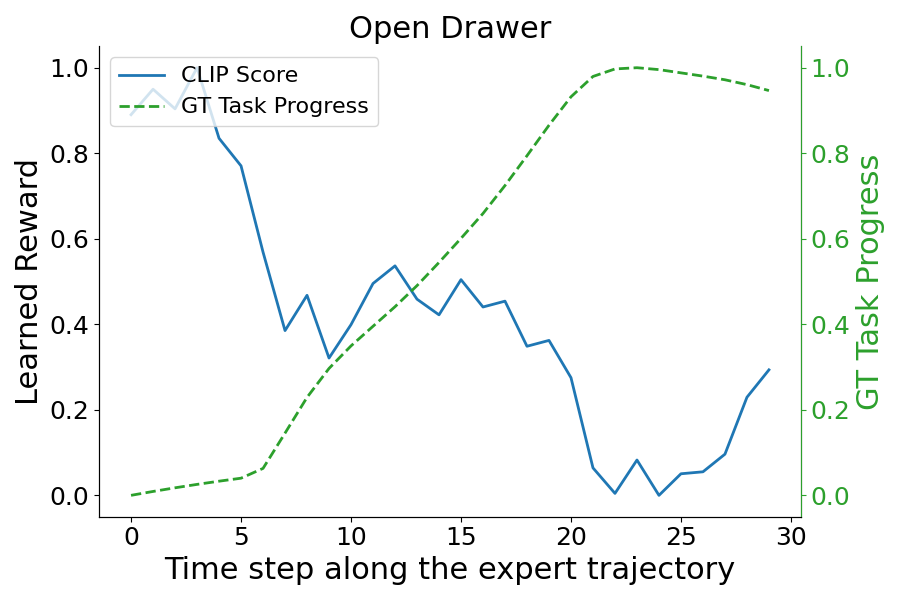}
    \includegraphics[width=.245\textwidth]{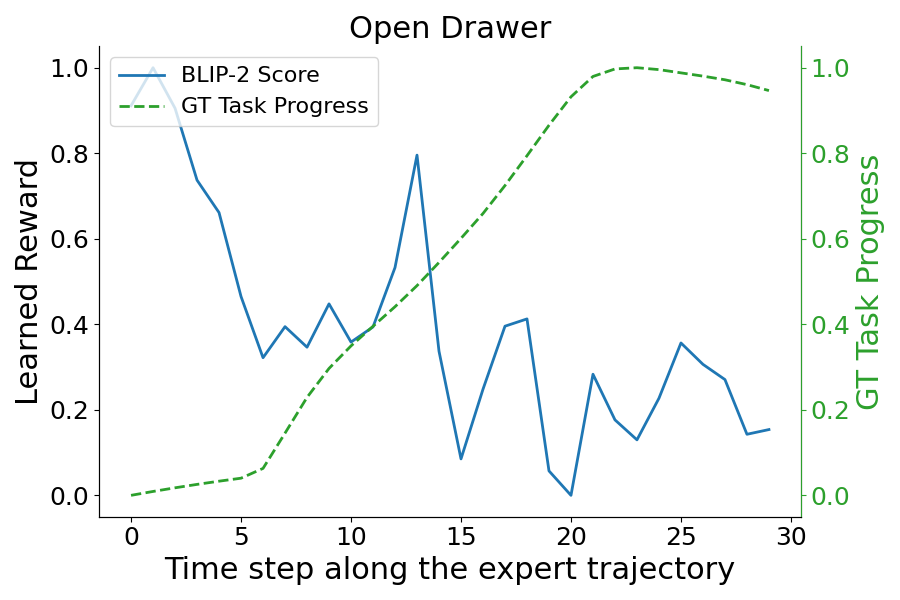}
    \\
        \includegraphics[width=.245\textwidth]{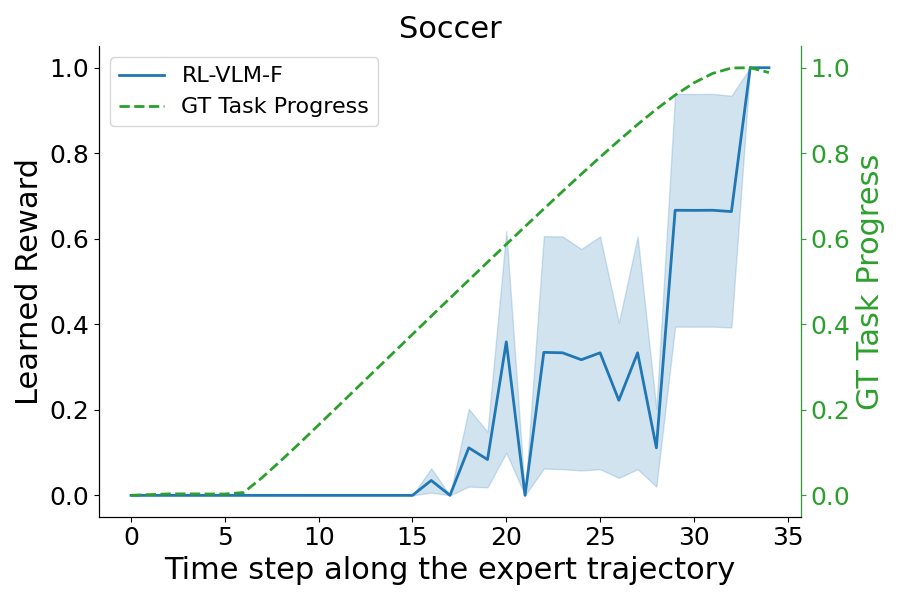}
    \includegraphics[width=.245\textwidth]{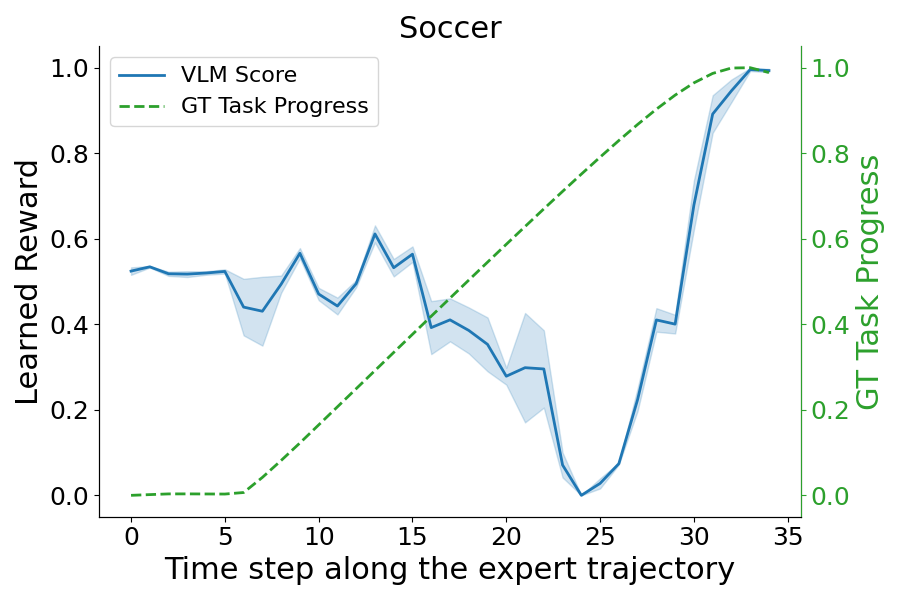}
    \includegraphics[width=.245\textwidth]{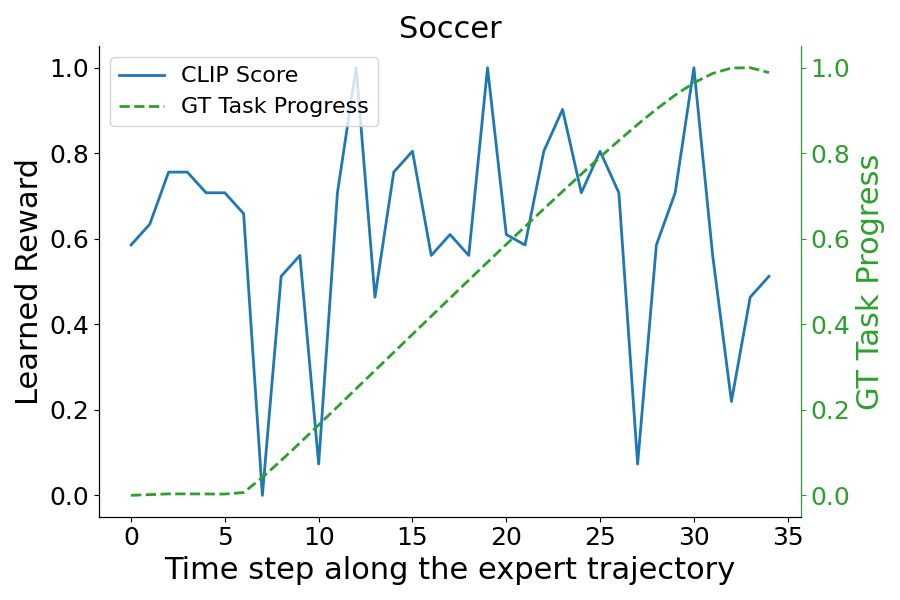}
    \includegraphics[width=.245\textwidth]{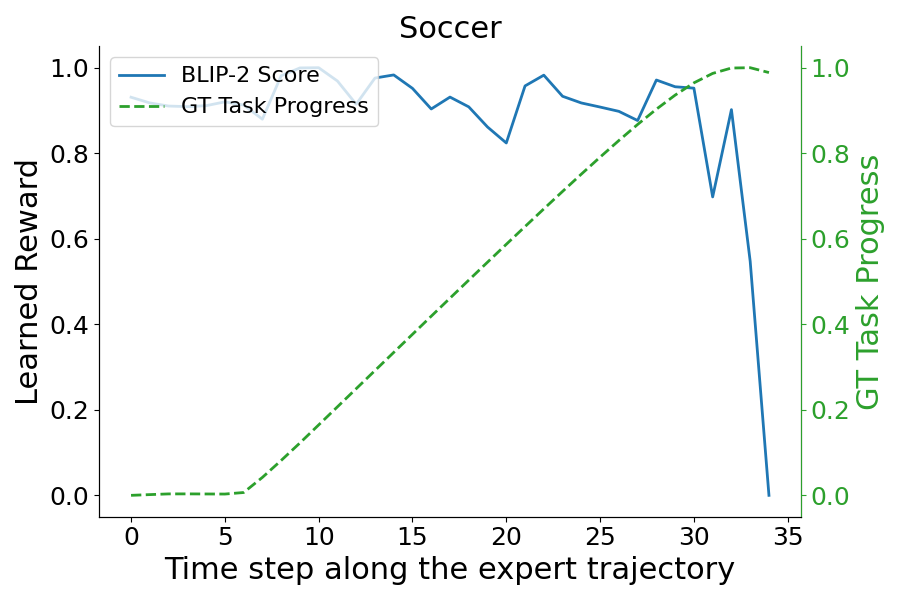}
    \\
     \includegraphics[width=.245\textwidth]{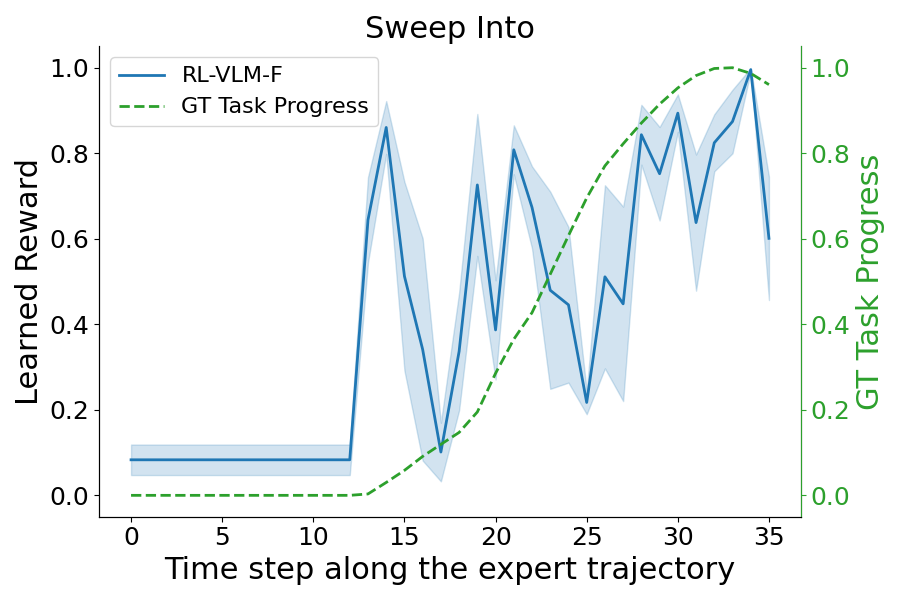}
    \includegraphics[width=.245\textwidth]{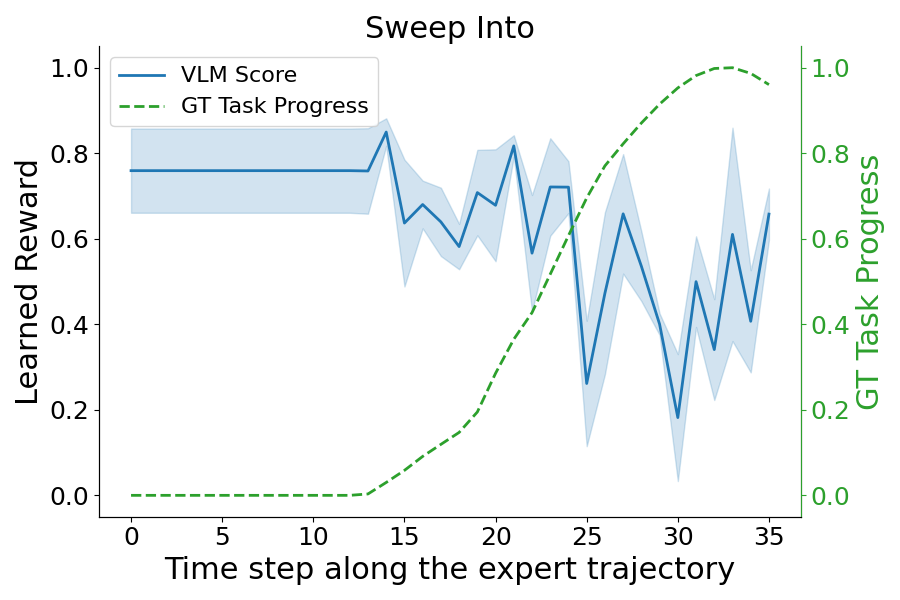}
    \includegraphics[width=.245\textwidth]{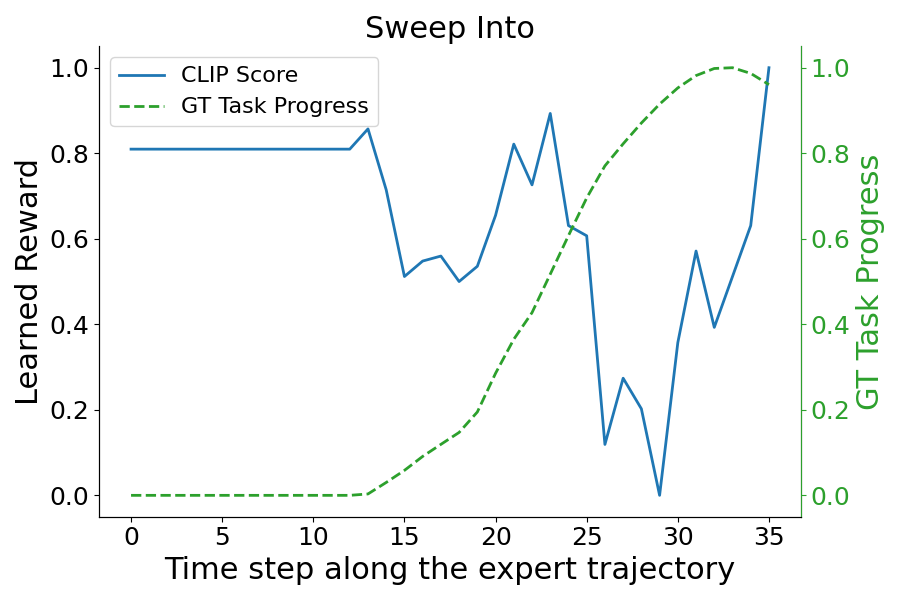}
    \includegraphics[width=.245\textwidth]{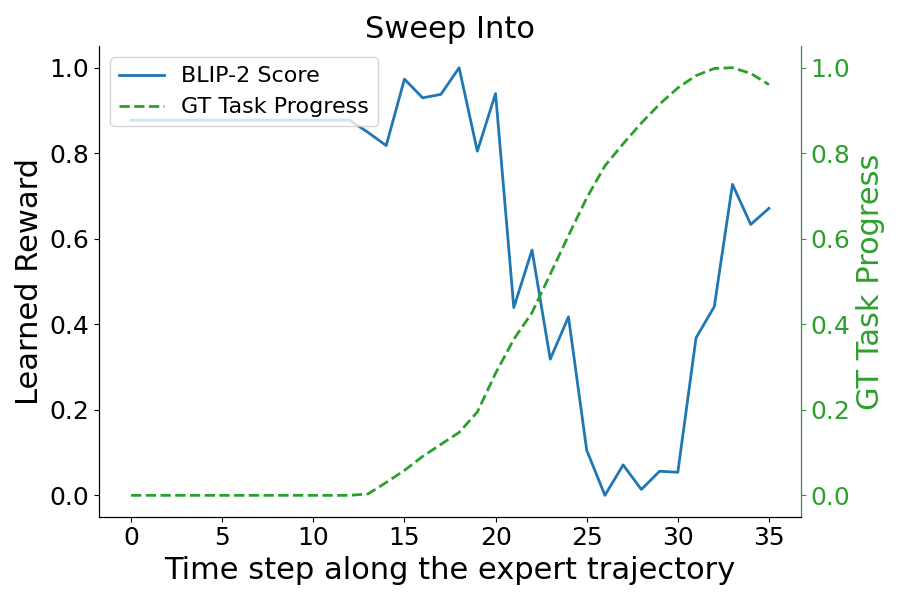}

    \caption{Comparison of learned reward functions from \method\ and VLM Score, as well as CLIP and BLIP-2 score to the ground-truth task progress along a trajectory rollout on three MetaWorld tasks. 
    From left column to right: reward learned by \method, reward learned by VLM Score, CLIP Score, BLIP-2 Score. From top row to bottom: \textit{Open Drawer}, \textit{Soccer}, and \textit{Sweep Into}. The reward learned by \method\ aligns the best across all compared methods. 
    }
    \label{fig:learned-reward}
\end{figure*}

Here we show the learned reward from \method\ and the VLM Score baseline, as well as the CLIP and BLIP-2 score along an expert trajectory on three MetaWorld tasks. We compare the learned reward from \method\ and the VLM Score / CLIP and BLIP-2 score to the ground-truth task progress. The results are shown in Figure~\ref{fig:learned-reward}. For all three tasks, the reward learned by \method\ aligns the best with the ground-truth task progress.

\bibliographystyle{icml2024}


\end{document}